\documentclass[lettersize,journal]{IEEEtran}
\usepackage{amsmath,amsfonts}
\usepackage{algorithm}
\usepackage{array}
\usepackage[caption=false,font=normalsize,labelfont=sf,textfont=sf]{subfig}
\usepackage{textcomp}
\usepackage{stfloats}
\usepackage{url}
\usepackage{verbatim}
\usepackage{graphicx}
\usepackage{cite}
\usepackage{multirow}
\usepackage{xcolor}
\usepackage{booktabs}

\usepackage[noend]{algpseudocode}
\algrenewcommand\algorithmicrequire{\textbf{Input:}}
\algrenewcommand\algorithmicensure{\textbf{Output:}}
\begin{document}

\title{LLM-Driven Instance-Specific Heuristic Generation and Selection}


\author{Shaofeng~Zhang, Shengcai~Liu, Ning~Lu, Jiahao~Wu, Ji~Liu, Yew-Soon~Ong, \and Ke~Tang%
\thanks{Shaofeng Zhang, Shengcai Liu and Ke Tang are with the Guangdong Provincial Key Laboratory of Brain-Inspired Intelligent Computation, Department of Computer Science and Engineering, Southern University of Science and Technology, Shenzhen 518055, and also with the Zhongguancun Academy, Beijing 100094. Email: 12445025@mail.sustech.edu.cn; liusc3@sustech.edu.cn; tangk3@sustech.edu.cn.
Ning Lu is with the Guangdong Provincial Key Laboratory of Brain-Inspired Intelligent Computation, Department of Computer Science and Engineering, Southern University of Science and Technology, Shenzhen 518055, and also with the Hong Kong University of Science and Technology, Hong Kong 999077;
Jiahao Wu is with the Guangdong Provincial Key Laboratory of Brain-Inspired Intelligent Computation, Department of Computer Science and Engineering, Southern University of Science and Technology, Shenzhen 518055, and also with the Hong Kong Polytechnic University, Hong Kong 999077.
Ji Liu is with the Advanced Micro Devices Inc., Beijing.
Yew-Soon Ong is with the Centre for Frontier AI Research, Institute of High Performance Computing, Agency for Science Technology and Research, Singapore 138632, and also with the College of Computing and Data Science, Nanyang Technological University, Singapore 639798. Email: asysong@ntu.edu.sg\\
\textbf{Acknowledgments:} This work was supported by the National Key Research and Development Program of China under Grant 2022YFA1004102, and in part by the Guangdong Major Project of Basic and Applied Basic Research under Grant 2023B0303000010, and the Zhongguancun Academy Project, (Grant No.s 02012403).}}



\markboth{IEEE Transactions on Emerging Topics in Computational Intelligence,~Vol.~XX, No.~X, XX~XXXX}%
{Zhang \MakeLowercase{\textit{et al.}}: LLM-Driven Instance-Specific Heuristic Generation and Selection}


\maketitle

\begin{abstract}
Combinatorial optimization problems are widely encountered in real-world applications.
A critical research challenge lies in designing high-quality heuristic algorithms that efficiently approximate optimal solutions within a reasonable time. 
In recent years, many works have explored integrating Large Language Models (LLMs) with Evolutionary Algorithms to automate heuristic algorithm design through prompt engineering. 
However, these approaches generally adopt a problem-specific paradigm, applying a single algorithm across all problem instances, failing to account for the heterogeneity across instances.
In this paper, we propose InstSpecHH, a novel framework that introduces the concept of instance-specific heuristic generation. InstSpecHH partitions the overall problem class into sub-classes based on instance features and performs differentiated, automated heuristic design for each problem subclass. By tailoring heuristics to the unique features of different sub-classes, InstSpecHH achieves better performance at the problem class level while avoiding redundant heuristic generation for similar instances, thus reducing computational overhead. This approach effectively balances the trade-off between the cost of automatic heuristic design and the quality of the obtained solutions.
To evaluate the performance of InstSpecHH, we conduct extensive experiments on 4,500 subclasses of the Online Bin Packing Problem (OBPP) and 365 subclasses of the Capacitated Vehicle Routing Problem (CVRP). Experimental results show that InstSpecHH demonstrates strong intra-subclass and inter-subclass generalization capabilities. Compared to previous problem-specific methods, InstSpecHH reduces the average optimality gap by 6.06\% for OBPP and 0.66\% for CVRP. These results highlight the potential of instance-aware automatic heuristic design to further enhance solution quality.
\end{abstract}

\begin{IEEEkeywords}
Automatic Algorithm Design, Code Generation, Large Language Models, Combinatorial Optimization.
\end{IEEEkeywords}

\section{Introduction}
Efficiently finding the solutions of NP-hard combinatorial optimization problems has long been a central research focus in computational intelligence~\cite{Bengio2021, LiLiu2024, WangYong2025}. 
Unlike exact algorithms, which require substantial computational resources as the problem scale increases~\cite{kool2018}, heuristic algorithms can generate high-quality solutions within a reasonable time, demonstrating considerable practical value in engineering applications.
However, traditional heuristic algorithm designs exhibit two critical limitations.
Firstly, their development heavily depends on human expertise, leading to prohibitively high design costs~\cite{burke2013}.
Secondly, manual design paradigms often inadequately explore the heuristic algorithm space, making it challenging to discover more effective heuristic strategy combinations~\cite{haoran2024reevo}.

The emergence of Large Language Models (LLMs) offers a viable approach to significantly reduce the cost of generating heuristics~\cite{LiuSurvey2026, TanKC2025arXiv250908269Z}, thereby enabling the exploration of a broader heuristic algorithm space.
Recently, several studies have attempted to leverage LLMs to automatically generate heuristics, aiming to utilize the strong representational and generalization capabilities of these models to assist in generating heuristic algorithms for solving NP-hard optimization problems~\cite{romera2024funsearch, fei2024eoh, haoran2024reevo}.
Specifically, these studies integrate LLMs with Evolutionary Algorithms (EAs), employing iterative querying and evaluation mechanisms to progressively generate the Python code of the heuristic algorithm tailored to a specific problem class.

However, such approaches often attempt to generate a single algorithm to solve all problem instances belonging to the class, without adequately accounting for the heterogeneity across different instances. In this paper, we refer to this paradigm as \textbf{LLM-driven problem-specific heuristic generation}. 
As a result, it often performs poorly over the entire problem class.
EoH~\cite{fei2024eoh} is one representative work in this line. 
Its original paper demonstrates that when provided with 5 test instances of the Online Bin Packing Problem (OBPP), EoH can generate a heuristic algorithm that outperforms Best Fit, a traditional heuristic for OBPP.
However, when we consider 94,500 instances with diverse problem features as test instances, the single heuristic generated by EoH achieves almost the same solution quality as Best Fit, as shown in Table~\ref{tab:intra_main}.
In such cases, it is difficult to obtain further improvements with a single heuristic algorithm.

To address the aforementioned limitation and achieve better performance, a straightforward approach is to leverage existing methods~\cite{fei2024eoh, haoran2024reevo} to generate a single heuristic for each individual instance within the problem class.
While this can improve performance, the cost is prohibitively high.
As illustrated in Table~\ref{tab:eoh_example}, even generating a single heuristic can require considerable computational effort. For instance, executing 800 queries with DeepSeek-V3 requires approximately 3 hours of computation time, making this approach impractical when faced with a large number of problem instances.
This naturally leads to a fundamental research question: \textit{How can we best account for instance heterogeneity under a limited computational budget}?

\begin{table}[]
\centering
\caption{\textbf{Single-run statistics of EoH under different settings}: Algorithm quality is measured by the ratio between required bins and the lower bound. The Code generation LLM is DeepSeek-V3.}
\label{tab:eoh_example}
\begin{tabular}{ccc}
\toprule
Settings                   & Case1 & Case2 \\ \midrule\midrule
\# Generation              & 20    & 16    \\
\# Population              & 20    & 10    \\
\# LLM Queries             & 2000  & 800   \\
Initial Quality (\%)       & 96.08 & 96.08 \\
Best Achieved Quality (\%) & 97.23 & 96.96 \\
Time / hour                & 6.86  & 2.85  \\ \bottomrule
\end{tabular}
\end{table}

In this work, we propose an \textbf{instance-specific heuristic generation and selection framework}, named InstSpecHH, to address the above question.
This framework assumes that problem instances with similar features can be effectively solved using the same heuristic algorithm.

The core process of InstSpecHH consists of an offline construction phase and an online inference phase. In the offline phase, the original instance space is partitioned into distinct subclasses based on problem features. Instances within each subclass share common characteristics, allowing them to be solved by a unified heuristic. Then, by integrating EAs with the LLM, a specific heuristic algorithm tailored to each subclass is generated. During the online phase, InstSpecHH leverages the LLM to select the most appropriate heuristic algorithm from the available problem subclasses to solve the target instance.
In this way, InstSpecHH enables adaptation to instance-level variation in an instance-specific manner, while avoiding the inefficiency of generating a separate heuristic for every individual instance. 
Although the first two steps, namely problem subclass construction and heuristic generation, are performed offline and typically incur relatively high computational cost, the final step is performed online, allowing the overall cost to be gradually amortized as the number of inference instances increases. Experimental results demonstrating this efficiency are presented in Section~\ref{sec:inv_efficiency}.

The key contributions of this work are outlined below:
\begin{itemize}
\item \textbf{Revealing Limitation of Problem-Specific Heuristic Generation}: This work identifies the limitation of using a single heuristic algorithm to solve an entire problem class, the failure to account for the heterogeneity among individual instances, which often leads to suboptimal solution quality.
\item \textbf{Pioneering Instance-Specific Framework with LLMs}: This work proposes InstSpecHH, the first LLM-driven instance-specific framework that partitions the problem class into subclasses to assist in heuristic algorithm generation and selection.

\item \textbf{Comprehensive Evaluation}: Extensive experiments conducted on 4,500 OBPP and 675 CVRP problem subclasses demonstrate that InstSpecHH achieves optimality gap reductions of over 6.06\% and 0.66\%, respectively, relative to previous problem-specific methods.
\end{itemize}

The remainder of this paper is organized as follows. Section~\ref{sec:related_work} introduces existing approaches to automated algorithm design and algorithm selection, and analyzes their limitations. Sections~\ref{sec:framework} and~\ref{sec:instantiation} describe the proposed framework and present framework instantiations for the Online Bin Packing Problem (OBPP) and Capacitated Vehicle Routing Problem (CVRP). Section~\ref{sec:experiments} provides the experimental analysis. Section~\ref{sec:conclusion} concludes the paper with a discussion of future work. The code and data associated with this work will be made publicly available.
\section{Related Work}\label{sec:related_work}
This section reviews the related literature from two methodological perspectives. We first examine traditional automated algorithm design and algorithm selection, neither of which incorporates LLMs. We subsequently review recent advances in LLM-based heuristic design and selection, while analyzing the limitations of existing approaches.

\subsection{Traditional Automated Algorithm Design}
Designing customized heuristics manually is highly labor-intensive. To automate this process and reduce reliance on human engineering, research has primarily advanced along two distinct paths: Automatic Algorithm Configuration and Neural Combinatorial Optimization.

Automatic Algorithm Configuration (AAC) aims to identify optimal parameter combinations that maximize the target algorithm's average performance across a set of representative instances. Existing AAC approaches are generally categorized into model-free and model-based methods~\cite{Schede2022}. Model-free methods, such as evolutionary algorithms~\cite{Carlos2009}, navigate the configuration space by directly evaluating the actual performance of candidate parameters. In contrast, model-based methods, such as Bayesian Optimization~\cite{Hutter2011}, construct surrogate models to predict parameter performance prior to execution, offering significantly higher sample efficiency under limited evaluation budgets~\cite{Amine2021}.

Moving beyond parameter configuration, Neural Combinatorial Optimization (NCO) offers a data-driven alternative by leveraging machine learning to solve NP-hard problems~\cite{Bengio2021, Mazyavkina2021}. From a heuristic design perspective, the neural network's architecture defines the heuristic algorithm space, and the training process systematically explores this space to optimize network parameters. Existing neural solvers can generally be categorized into two main classes: constructive methods and improvement methods. Constructive methods aim to learn a policy that incrementally builds a complete solution through a sequence of decisions, such as Pointer Networks~\cite{Vinyals2015PointerNet} and reinforcement-learning-guided Transformers~\cite{Bello2016, kool2018}. In contrast, improvement methods, often referred to as Learning Improvement Heuristics (LIH), focus on iteratively refining an existing initial solution~\cite{Chen2019, Ye2025}. However, despite these advancements in automation, these paradigms remain fundamentally constrained by human-designed algorithmic or neural architectures.

\subsection{Algorithm Selection and Instance-Specific Configuration}
The fundamental motivation for instance-specific algorithm configuration is rooted in the No Free Lunch (NFL) theorem~\cite{Wolpert1997}, which mathematically demonstrates that no ``universal'' algorithm dominates across all problem instances. To address this limitation within a single solver, Adaptive Metaheuristics dynamically adjust internal parameters or search operators during execution~\cite{Pei2025}. For example, Adaptive Large Neighborhood Search (ALNS) autonomously balances exploration and exploitation via continuous feedback~\cite{Ropke2006, Keskin2016}. To overcome the limitations of traditional weight-based scoring, recent advancements formulate ALNS operator selection as a Markov Decision Process. By embedding deep Q-networks (DQNs), these data-driven approaches extract real-time search states to adaptively select optimal heuristic combinations (e.g., destroy, repair, and charging strategies) for complex routing problems~\cite{Wang2025}.

Rather than adapting a single solver, Automated Algorithm Selection (AAS) addresses heterogeneity by routing specific instances to the best available solver. AAS typically maps instance features to performance using either classification (predicting the best algorithm's ID) or regression (predicting specific metrics for finer distinction)~\cite{Kotthoff2016}. In addition, Wu et al. proposed a DAG-based algorithm selection framework~~\cite{wu2025towards} that models the distribution of algorithm features conditioned on problem features, thereby improving robustness under distribution shift while providing both model-level and instance-level explainability. Extending this instance-aware paradigm, frameworks like ISAC~\cite{ISAC} adaptively cluster instances into sub-classes, assigning optimal parameter configurations to each. To eliminate manual feature engineering, recent deep learning approaches~\cite{Zhao2021} map raw instances directly into a latent space via encoders, achieving fully automated, feature-free algorithm selection.

Despite addressing instance heterogeneity, these frameworks remain constrained by small, human-designed algorithm libraries. High design costs limit candidate variety, preventing fine-grained selection tailored to the specific nuances of individual problem instances.

\begin{figure*}[t]
    \centering
    \includegraphics[width=0.8\textwidth]{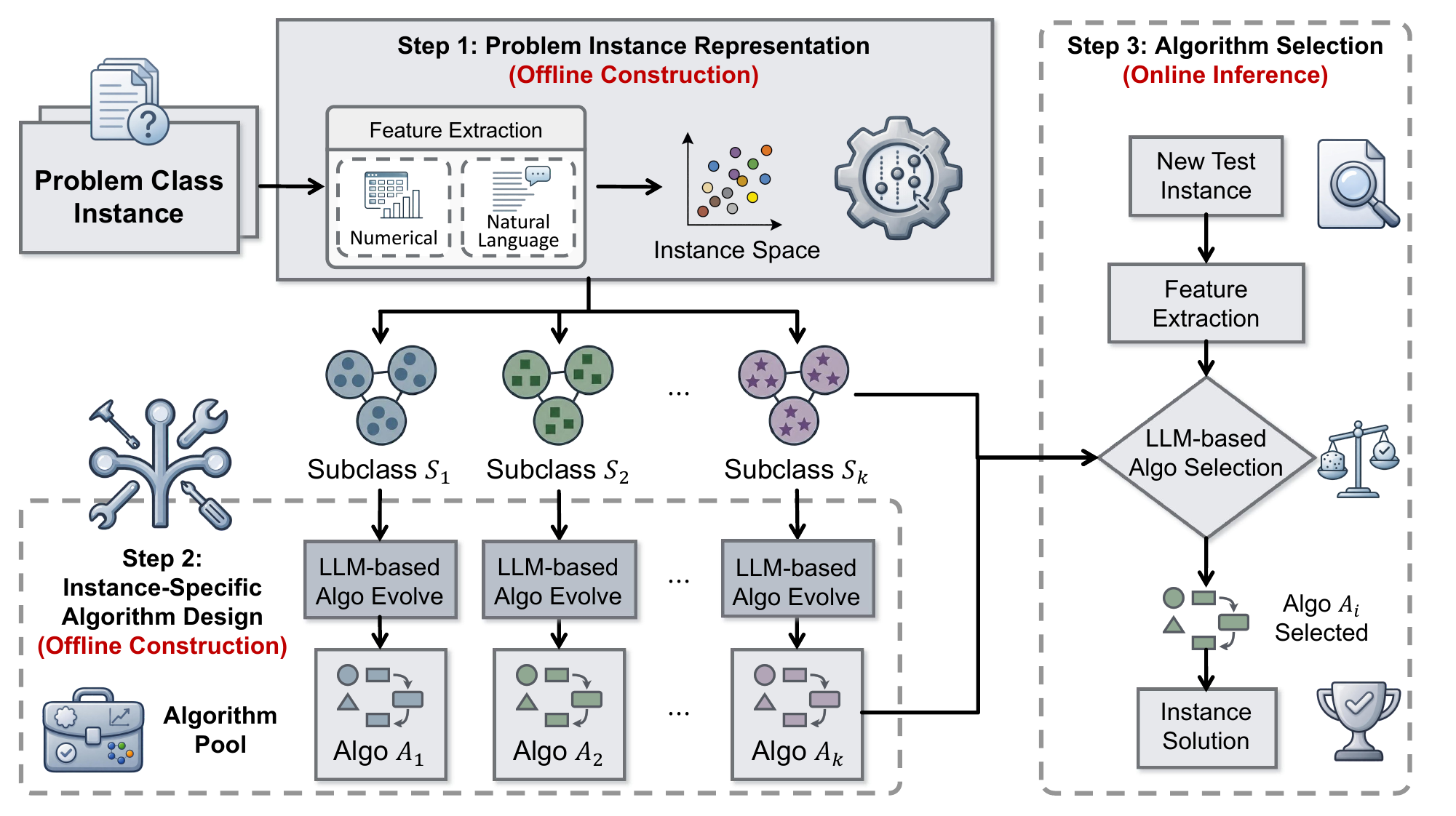}
    \caption{InstSpecHH Diagram. First, construct problem subclasses and their representations (top). Next, generate instance-specific heuristic algorithms (middle). Finally, select the best‐matching heuristic algorithm for the target instance (bottom).}
    \label{fig:workflow}
\end{figure*}

\subsection{LLM-Based Heuristic Design and Selection}
Recent advances in LLMs have achieved remarkable success across a wide range of tasks~\cite{LiuCQ0O24}. Particularly, their robust code-generation and zero-shot reasoning capabilities have sparked a novel paradigm in automated algorithm design.
The paradigm of algorithm design via LLMs focuses on generating high-quality heuristics for complex optimization problems through iterative search, typically by integrating LLMs with Evolutionary Algorithms (EAs)~\cite{WuXingyu2025, LLaMEA, Liu2024LLM4AD}. Representative frameworks like FunSearch~\cite{romera2024funsearch} utilize evolutionary strategies to search the function space, pairing an LLM that mutates code snippets with an automated evaluator. EoH~\cite{fei2024eoh} further refines this population-based approach by explicitly prompting LLMs to execute selection, crossover, and mutation operations, effectively fusing parental strengths. Building on this, ReEvo~\cite{haoran2024reevo} introduces a cognitive ``reflection'' mechanism, analyzing historical execution data to produce ``verbal gradients'' that explicitly guide the LLM toward superior code. To further refine this search process and balance exploration and exploitation, HSEvo~\cite{Dat_Doan_Binh_2025} introduces novel diversity metrics that improve heuristic performance at low computational costs. Concurrently, MCTS-AHD~\cite{zheng2025monte} innovatively incorporates Monte Carlo Tree Search to mitigate premature convergence, enabling a deeper exploration of the heuristic space. These enhanced generative capabilities have driven successful applications in diverse domains~\cite{LiuSurvey2026}.

While these LLM-driven generation frameworks represent significant breakthroughs, they predominantly adopt a problem-specific paradigm, aiming to discover a single global heuristic for an entire problem class. Consequently, they overlook the inherent heterogeneity across diverse problem instances. Recognizing the necessity of instance awareness, recent studies have introduced LLM-enhanced algorithm selection~\cite{Wu2024}, which automatically extracts comprehensive structural and semantic representations directly from code text to align algorithm features with problem representations for accurate selection. However, akin to classical algorithm selection, this paradigm merely routes instances among existing options and remains strictly bottlenecked by predefined algorithm portfolios, typically limited to a small pool of fewer than 50 candidate algorithms. Besides, it lacks flexibility: any modification, update, or addition to the algorithm pool necessitates a complete and costly retraining of the selection model.

\section{The InstSpecHH Framework}\label{sec:framework}

Unlike traditional problem-specific paradigms that rely on a single heuristic to solve all instances within a problem class, this study considers the heterogeneity among problem instances and introduces an automatic algorithm design and selection framework, referred to as \textbf{InstSpecHH}, as illustrated in Figure~\ref{fig:workflow}.
In this framework, problem instance representation and heuristic algorithm design are conducted offline, while heuristic algorithm selection is performed online. The three components are detailed as follows:
\begin{itemize}
    \item \textbf{Problem Instance Representation}: The entire problem class is partitioned into multiple subclasses. Each subclass is described using key features, providing a foundation for instance-specific algorithm design and selection.
    \item \textbf{Heuristic Algorithm Design}: For each identified problem subclass, InstSpecHH automatically generates instance-specific heuristic algorithms tailored to the corresponding subclass.
    \item \textbf{Heuristic Algorithm Selection}: Based on the features of a given problem instance, InstSpecHH matches it with the most relevant problem subclass and selects the most suitable existing algorithm accordingly.
\end{itemize}

\subsection{Problem Instance Representation} \label{sec:subclass_representation}
A problem instance refers to a specific input (data) for a given problem, containing all the necessary information to solve it. Since problem instances often share similar features, repeatedly designing heuristic algorithms would result in significant redundancy and unnecessary costs. 
Thus, it is essential to group these problem instances into subclasses to reduce unnecessary overhead in heuristic algorithm design and to provide key information for algorithm selection.
Each group of instances sharing the same features is referred to as a \textbf{problem subclass}, and for each subclass, a corresponding representation is constructed, including both feature-based and natural language-based representations.

\begin{figure*}[t]
    \centering
    \includegraphics[width=1\textwidth]{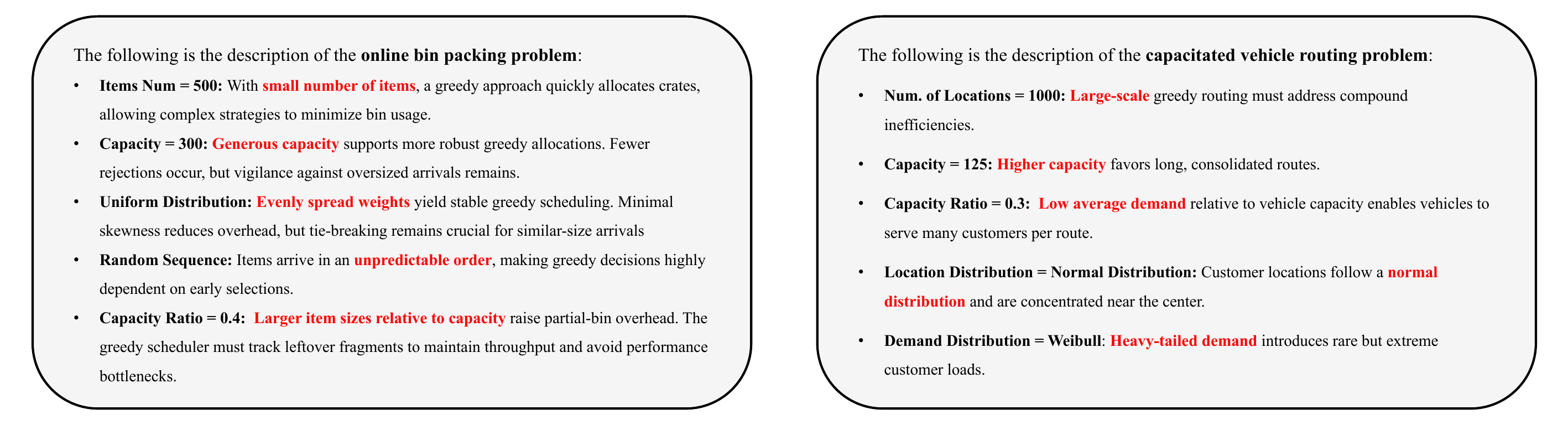}
    \caption{Example of natural language description for OBPP (left) and CVRP (right) problem features.}
    \label{fig:prompt}
\end{figure*}

To group problem instances, we design a set of key problem features that capture the structural and statistical characteristics relevant to heuristic performance.
Each problem feature consists of multiple discrete feature values, allowing problem instances to be distinctly identified and grouped accordingly. 
Let $i$ denote a specific problem instance characterized by $d$ distinct features. 
The value of the $m$-th feature of instance $i$ is represented as $x_i^m$. 
When certain features are categorical rather than numerical, they are encoded as integer IDs. Consequently, the feature information of problem instance $i$ can be represented by a feature vector (i.e., feature-based representations):
\begin{equation}
    \mathbf{x}_i = (x_i^1, x_i^2, \dots, x_i^d) \in \mathbb{R}^d.
    \label{eq:feature_vector}
\end{equation}
When two instances share an identical feature vector $\mathbf{x}_i$, they are categorized into the same subclass. Furthermore, through interaction with LLMs, we generate descriptions corresponding to each feature value $x_i^m$ to aid in the characterization of problem subclasses. These descriptions serve as language-based representations of the problem instance, as illustrated in Figure~\ref{fig:prompt}. Additionally, Section~\ref{sec:instantiation} demonstrates two example optimization problems to illustrate the proposed problem representation.

After establishing the feature values for each feature, we can partition the entire problem class accordingly. 
Suppose that the $m$-th feature has $|x^m|$ distinct possible values; then, a total of $|x^1| \times |x^2| \times \dots \times |x^d|$ problem subclasses can be constructed. 
We design instance generators tailored to each specific problem subclass, ensuring that the generated test instances are consistent with the corresponding feature values.

\begin{algorithm}[htbp]
\caption{Heuristic Algorithm Design}
\label{alg: had}
\begin{algorithmic}[1]          
\Require Problem pool $\mathcal{I}$, evolutionary operator pool $\mathcal{O}$, evaluate function $f(\cdot)$, population size $N$, maximum generation $G_{\max}$, problem distance function $d(\cdot)$, LM $LM_\theta$, number of neighbors in NS $k_n$
\Ensure  algorithm pool $\mathcal{H}$

\State // Generate the heuristic algorithm for each problem subclass instance
\ForAll{problem subclass $s \in$ problem pool $\mathcal{I}$}
    \State $t_s \gets LM_\theta (s, \text{Desc})$ \Comment{Instance-specific prompt}
    \State $P_0 \gets \{\,h_1, h_2, \dots, h_N\,\}$ \Comment{Initialize population} \label{line: init}
    \State Evaluate fitness $f(h, s)$ for every individual heuristic in $P_0$ \label{line: eval}
    \For{$g \gets 1$ \textbf{to} $G_{\max}$}
        \State // Repeat Line~\ref{line: evo}-\ref{line: update} with different operator $o$
        \State Get evolutionary prompt $t_e$ with operator $o\in \mathcal{O}$ \label{line: evo}
        \State Choose parents $M_g$ from $P_{g-1}$
        \State Generate offspring $C_g \gets LM_\theta (t_p, t_e, M_g)$
        \State Evaluate the fitness of offspring $C_g$
        \State Update $P_{g-1}$ with $C_g$, yielding the $P_g$ \label{line: update}
    \EndFor

    \State // Select the algorithm with the highest score
    \State $\mathcal{H}[s] \gets \displaystyle\arg\max_{s \in \bigcup_{g=0}^{G_{\max}} P_g} f(h,s)$
\EndFor

\State // Neighbor search
\ForAll{$s \in \mathcal{I}$}
  \State Compute distance $d(s,s')$ to $s$ for every $s' \in \mathcal{I}$ 
  \State $D_s^k \gets$ the $k_n$ closest subsets to $s$ \label{line: closest}
  \State $\mathcal{H}[s] \gets \displaystyle\arg\max_{s' \in D_s^k} f({H}[s'],s)$ \label{line: select}
\EndFor
\State \Return $\mathcal{H}$
\end{algorithmic}
\end{algorithm}

\subsection{Heuristic Algorithm Design}
Unlike previous work~\cite{romera2024funsearch,fei2024eoh,haoran2024reevo}, which generates a single heuristic algorithm for the entire problem class, this section aims to develop heterogeneous heuristic algorithms tailored to each problem subclass in an offline manner. Building on the subclass definitions introduced in Section~\ref{sec:subclass_representation}, and inspired by the approach in~\cite{fei2024eoh}, we generate instance-specific heuristic algorithms by combining LLMs with EAs, as shown in Algorithm~\ref{alg: had}.

For each problem subclass, the population $P$ is initially established (Line~\ref{line: init}), where each $h$ corresponds to a candidate heuristic algorithm. The performance of each candidate heuristic is then evaluated using test instances representative of the respective problem subclass (Line~\ref{line: eval}).
Various genetic operators, such as crossover and mutation, are applied in sequence during each evolutionary iteration. For each operator, the corresponding prompt is generated. Next, parent heuristic algorithms $M_g$ are chosen via roulette wheel selection, and the LLM is used to produce the operation-specific offspring $C$. The newly generated algorithm $C_g$ is subsequently evaluated using test instances from the targeted subclass. Its performance quality is quantified as a fitness score, which guides the updating of the heuristic population $P_{g-1}$. Candidates demonstrating superior fitness scores are retained, while lower-performing heuristics are systematically phased out from the population (Line~\ref{line: update}).
This iterative evolutionary refinement optimizes the heuristic algorithm population, leading to progressively more robust and customized heuristic solvers tailored to each problem subclass. 
The final InstSpecHH algorithms are the heuristic algorithm pool constructed for each subclass. We have provided two examples in the Supplementary for the OBPP and CVRP problems, respectively.

\begin{figure*}[t]
    \centering
    \includegraphics[width=0.8\textwidth]{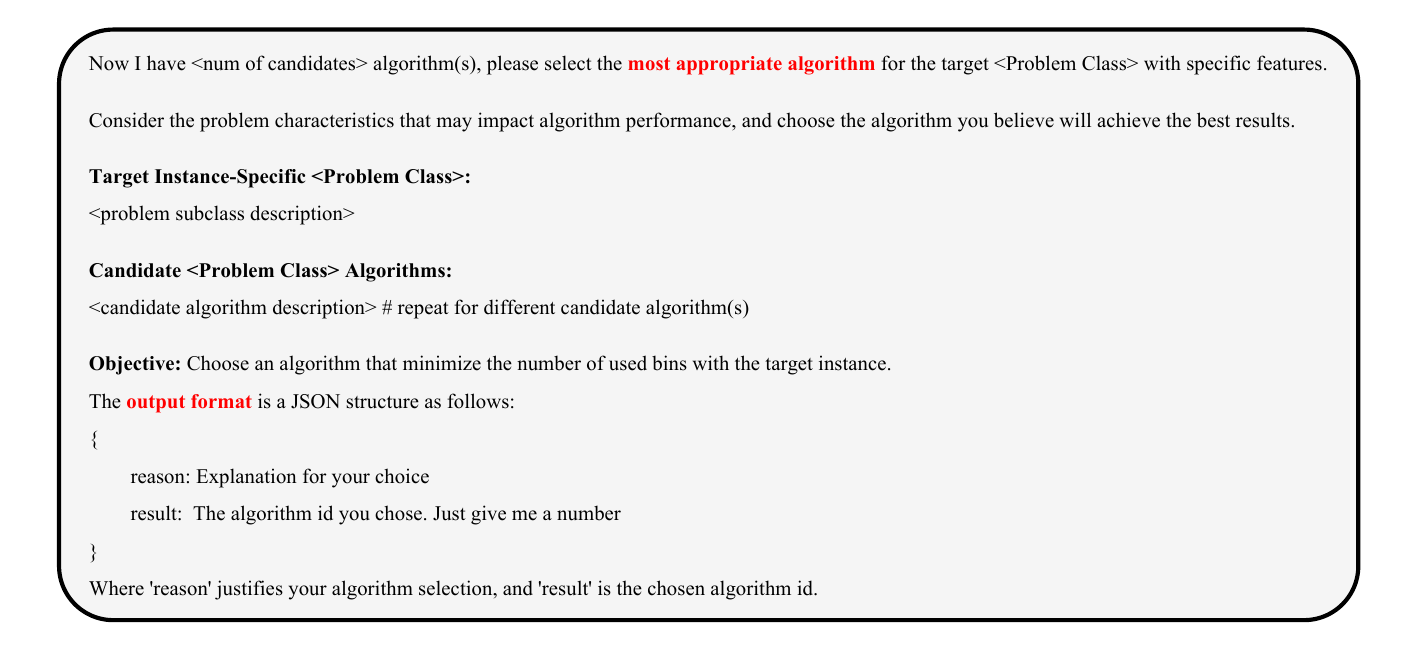}
    \vskip -0.1in
    \caption{Heuristic Algorithm Selection Prompt Template.}
    \label{fig:algo_selection}
\end{figure*}

In our experiments, we observed a tendency for the population-based evolutionary process to become constrained by local optima. To mitigate this limitation and further advance the quality of automatically designed algorithms, we introduced a \textbf{neighbor search (NS)} strategy. This strategy assumes that closely related problem subclasses exhibit similar instance distributions and thus possess potentially transferable solution strategies. Specifically, for each target problem subclass, we identify and retrieve the $k_n$ closest subclasses, referred to as neighbors, and incorporate their existing heuristic algorithms as additional candidate solutions (Line~\ref{line: closest}). These candidate algorithms undergo evaluation with the target subclass, with the top-performing heuristic selected as the final solver for the target subclass (Line~\ref{line: select}). Incorporating neighbor search effectively broadens the search space and substantially enhances the overall solution quality produced by the automated design framework.

By designing instance-specific heuristic algorithms for various problem subclasses, a diverse heuristic algorithm pool can be constructed, which provides richer candidate solutions for different problem instances and enhances the robustness and generalization capability of optimization problem solving.

\subsection{Heuristic Algorithm Selection}\label{sec:algo_selection}
Motivated by the impressive reasoning and expressive capabilities of LLMs in natural language processing~\cite{feng2025prm}, this study leverages LLMs to assist in online selection of the optimal heuristic algorithm for specific problem instances. This paradigm shift raises a critical question: \textit{Are LLMs inherently better suited for algorithm selection than traditional Neural Networks?}

To prevent context overflow for the LLM and reduce classification complexity for the neural network, a distance-based pre-selection mechanism first identifies candidate algorithms. For a target instance $i$, its standardized feature vector $\tilde{\mathbf{x}}_i$ is computed, and the $k_c$ closest problem subclasses are retrieved based on Euclidean distance:
\begin{equation}
    d(i, i') = \|\tilde{\mathbf{x}}_i - \tilde{\mathbf{x}}_{i'}\|_2.
    \label{eq:dist}
\end{equation}
By measuring similarity exclusively in the instance feature space, this mechanism decouples the algorithm selection from the heuristic pool, ensuring the selection process remains invariant to any structural or code-level modifications of the available heuristics.
The heuristics associated with this top-$k_c$ set, denoted as $\mathcal{N}_{k_c}(i)$, form the refined candidate pool. Subsequently, algorithm selection is executed via two distinct strategies:

\textbf{LLM-Based Selection:} The target instance's feature vector and the textual profiles of the candidate algorithms in $\mathcal{N}_{k_c}(i)$ are translated into a natural-language prompt (Figure~\ref{fig:algo_selection}). Guided by this context, the LLM acts as a reasoning agent, assessing the feature-algorithm fit to logically deduce the optimal heuristic.

\textbf{Neural Network Classifier:} Alternatively, a lightweight two-layer feedforward network $f_{\theta}(\tilde{\mathbf{x}})$ is trained on the intra-subclass dataset using cross-entropy loss. To ensure robust routing and align with the distance-based pre-selection, the classifier's output probabilities $\tilde{p}_c(\tilde{\mathbf{x}})$ are masked, restricting the final selection strictly to the pre-selected candidates:

\begin{equation}
    a^*=\arg\max_{a\in\mathcal{N}_{k_c}(i)}\ \tilde{p}_c(\tilde{\mathbf{x}}).
\end{equation}

\textbf{LLM vs. Neural Network:} These approaches represent fundamentally distinct paradigms for heuristic selection. While the NN classifier facilitates rapid inference, it remains bottlenecked by a fixed algorithm pool; any additions or modifications to the heuristic pool require costly retraining. In contrast, the LLM-based approach naturally overcomes these scalability limitations. It seamlessly adapts to dynamic, evolving algorithm libraries without requiring weight updates, while simultaneously offering interpretable reasoning for its routing decisions. Ultimately, this establishes a more flexible, generalized, and insightful paradigm for automated algorithm selection. 
We further perform a comparative experimental study of the two approaches. The results in Section~\ref{sec:inter_gen} indicate that they achieve comparable overall performance, suggesting that LLM-based algorithm selection is a promising alternative to conventional neural network classifiers. However, when the number of candidate algorithms becomes larger, the neural network classifier demonstrates stronger robustness, indicating a limitation of the LLM-based approach in such settings, as shown in Section~\ref{sec:sen_candidate}.

\section{Framework Instantiation for OBPP and CVRP}\label{sec:instantiation}

To instantiate the proposed framework, this section focuses on two representative optimization problems: the Online Bin Packing Problem (OBPP) and the Capacitated Vehicle Routing Problem (CVRP). To support instance-specific algorithm selection, problem classes are partitioned into subclasses based on features that significantly influence heuristic performance, specifically capturing problem scale, constraints, and data distributional properties.

\textbf{Online Bin Packing Problem (OBPP).} In OBPP, items of varying sizes arrive sequentially and must be irrevocably packed into bins of fixed capacity. The objective is to minimize the total number of bins used without exceeding capacity limits. The problem space is systematically partitioned by enumerating combinations of five key features:
\begin{itemize}
\item \textbf{Number of Items}: Specifies the instance scale, ranging from 500 to 5000 items.
\item \textbf{Item Weight Distribution}: Models the statistical characteristics of item weights (Uniform, Gaussian, or Weibull).
\item \textbf{Sequence Type}: Describes the arrival pattern of items (random, non-decreasing, or non-increasing by weight).
\item \textbf{Capacity}: Determines the size constraint for each bin (values from 50 to 500).
\item \textbf{Capacity Ratio}: Represents packing tightness, defined as the average item weight divided by the bin capacity (values in $[0.3, 0.7]$).
\end{itemize}

\textbf{Capacitated Vehicle Routing Problem (CVRP).} The objective of CVRP is to determine optimal routes for a fleet of uniform-capacity vehicles to serve geographically distributed customers exactly once, minimizing total travel distance without exceeding vehicle capacities. The CVRP domain is similarly partitioned using five key features:
\begin{itemize}
\item \textbf{Number of Customers}: Specifies the instance scale, ranging from 200 to 1000.
\item \textbf{Location Distribution}: Models the spatial layout of customers (Uniform, Gaussian, or Grid-based).
\item \textbf{Demand Distribution}: Captures the statistical variability in customer demands (Uniform, Gaussian, or Weibull).
\item \textbf{Vehicle Capacity}: Determines the load limit for each vehicle, selected from $\{50, 75, 100, 125, 150\}$.
\item \textbf{Capacity Ratio}: Reflects the overall instance tightness, defined as the ratio of average customer demand to vehicle capacity (values in $\{0.3, 0.5, 0.7\}$).
\end{itemize}

For both problems, an instance is characterized by a feature vector comprising these specific attributes. While enumerating these combinations systematically covers the predefined problem subclasses, out-of-distribution instances with unseen feature configurations are dynamically managed using the algorithm selection strategies introduced in Section~\ref{sec:algo_selection}.
\section{Experiments}\label{sec:experiments}
To systematically evaluate the proposed framework, we conduct comprehensive experiments to address the following research questions (RQs):

\begin{description}
\item[\textit{RQ1}:] Can instance-specific heuristic design achieve better performance by taking instance heterogeneity into account?
\item[\textit{RQ2}:] How does the performance of the LLM-based selection strategy compare with that of a classifier-based strategy?
\item[\textit{RQ3}:] How does InstSpecHH compare with the strong baseline that constructs a tailored heuristic algorithm for each individual problem instance?
\item[\textit{RQ4}:] How sensitive is InstSpecHH to its parameter settings?
\item[\textit{RQ5}:] Under which instance feature characteristics does the instance-specific paradigm significantly outperform the problem-specific paradigm?

\end{description}

\begin{table}[t]
\centering
\caption{Experimental Settings of InstSpecHH for OBPP and CVRP}
\label{tab:setting}
\renewcommand{\arraystretch}{1.2}
\setlength{\tabcolsep}{5pt}
\small
\begin{tabular}{llcc}
\toprule
\textbf{Category} & \textbf{Setting} & \textbf{OBPP} & \textbf{CVRP} \\
\midrule
\multirow{3}{*}{\textbf{Problem}}
 & \# instance & 30 & 30 \\
 & \# intra-subclass & 3,150 & 472 \\
 & \# inter-subclass & 1,350 & 203 \\
\midrule
\multirow{4}{*}{\textbf{Design}}
 & \# Population & 10 & 10 \\
 & \# LLM Queries & 800 & 800 \\
 & \# Neighbor Search $k_n$ & 20 & 20 \\
\midrule
\multirow{5}{*}{\textbf{Selection}}
 & LLM $k_c$ & 3 & 2 \\
 & Classifier $k_c$ & 5 & 2 \\
 & Classifier Hidden dim & 128 & 128 \\
\bottomrule
\end{tabular}
\vspace{2pt}
\end{table}

\begin{table*}[htbp]
\centering
\caption{\textbf{Intra-Subclass Generalization:} The test instances share the same problem feature combinations as the training instances. In OBPP, the objective value represents the number of bins used, while in CVRP it corresponds to the total vehicle travel distance. The optimality gap (Opt.\ Gap) is computed with respect to the best-known solution.}
\label{tab:intra_main}
\renewcommand{\arraystretch}{1.2}
\setlength{\tabcolsep}{8pt}
\begin{tabular}{l l c c}
\toprule
\textbf{Problem Class} & \textbf{Method} & \textbf{Obj. ($\downarrow$)} & \textbf{Opt. Gap \% ($\downarrow$)} \\
\midrule
\multirow{6}{*}{OBPP} 
 & First Fit & 1386.58 & 11.16 \\
 & Best Fit & 1383.78 & 10.94 \\
 & EoH & 1383.68 & 10.93 \\
 & ReEvo & 1383.78 & 10.94 \\
\cmidrule(lr){2-4}
 & InstSpecHH w/o NS & 1334.13 & 7.01 \\
 & InstSpecHH & \textbf{1304.47} & \textbf{4.67} \\
\midrule
\multirow{5}{*}{CVRP} 
 & Closest Priority & 450.22 & 18.59 \\
 & EoH & 405.12 & 5.11 \\
 & ReEvo & 407.63 & 6.13 \\
\cmidrule(lr){2-4}
 & InstSpecHH w/o NS & 413.85 & 7.72 \\
 & InstSpecHH & \textbf{400.95} & \textbf{4.45} \\
\bottomrule
\end{tabular}
\vspace{4pt}
\end{table*}

\subsection{Experimental Setting}
As introduced in Section~\ref{sec:instantiation}, the proposed framework is evaluated on OBPP (4,500 subclasses) and CVRP (675 subclasses). These subclasses are randomly split into intra- and inter-sets at a 7:3 ratio. For each intra-subclass, InstSpecHH generates a tailored heuristic using 30 training instances. Generalization capabilities are assessed across two dimensions: intra-subclass generalization (testing on 30 instances with seen feature combinations) and inter-subclass generalization (testing on 30 instances with unseen feature combinations). Detailed configurations are summarized in Table~\ref{tab:setting}.

The proposed framework is compared against manually designed heuristics (Best Fit and First Fit for OBPP; Closest Priority for CVRP) and two representative LLM-driven baselines: EoH~\cite{fei2024eoh} and ReEvo~\cite{haoran2024reevo}. Unlike InstSpecHH, these baselines ignore instance heterogeneity and generate a single, general-purpose heuristic per problem class. For all automated methods, the population size is set to 10, with a cap of 800 LLM queries. While EoH and ReEvo utilize DeepSeek-V3~\cite{2024DeepSeekv3}, InstSpecHH employs the more efficient DeepSeek-R1-Distill-Qwen-14B~\cite{deepseekr1} to manage the computational cost of generating instance-specific algorithms. The algorithm selection process is repeated 5 times.

Performance is evaluated using the objective value (Obj.), defined as the number of used bins for OBPP and the total travel distance for CVRP, alongside the optimality gap (Opt. gap) relative to a reference solution ($\text{Obj}_{\text{ref}}$):
\begin{equation}
\text{Optimality Gap} = \frac{\text{Obj}_{\text{alg}} - \text{Obj}_{\text{ref}}}{|\text{Obj}_{\text{ref}}|} \times 100\%
\end{equation}
where $\text{Obj}_{\text{alg}}$ denotes the objective value achieved by the evaluated algorithm. To establish $\text{Obj}_{\text{ref}}$, OBPP utilizes the theoretical lower bound (total item weight divided by bin capacity), whereas CVRP employs the baseline solution generated by the Hybrid Genetic Search (HGS) algorithm~\cite{HGS}. For both metrics, lower values indicate higher solution quality achieved by the algorithm.

All experiments were conducted on a server running Ubuntu 22.04.5 LTS, featuring two Intel(R) Xeon(R) Platinum 8378A CPUs (3.00 GHz, 128 threads in total), 1.0 TiB of RAM, and eight NVIDIA A800-SXM4-40GB GPUs.

\begin{table*}[htbp]
\centering
\caption{\textbf{Inter-Subclass Generalization}: The test instances exhibit distinct feature combinations from the training instances. \# LLM Queries denotes the average number of LLM query attempts. For LLM-based algorithm selection, the number of candidate heuristics is $k_c = 3$ for OBPP and $k_c = 2$ for CVRP. For classifier-based algorithm selection, the number of candidate heuristics is $k_c = 5$ for OBPP and $k_c = 2$ for CVRP.}
\label{tab:inter_main}
\renewcommand{\arraystretch}{1.2}
\setlength{\tabcolsep}{8pt}
\begin{tabular}{l l c c c}
\toprule
\textbf{Problem Class} & \textbf{Method} & \textbf{Obj. ($\downarrow$)} & \textbf{Opt. Gap \% ($\downarrow$)} & \textbf{\# LLM Queries ($\downarrow$)} \\
\midrule
\multirow{8}{*}{\centering OBPP} 
 & First Fit & 1345.94 & 11.07 & - \\
 & Best Fit & 1343.15 & 10.85 & - \\
 & EoH & 1343.05 & 10.84 & - \\
 & ReEvo & 1343.15 & 10.85 & - \\
\cmidrule(lr){2-5}
 & InstSpecHH w/o NS + LLM & 1290.66 $\pm$ 0.60 & 7.03 $\pm$ 0.01 & 1.01 \\
 & InstSpecHH + Random & 1266.26 $\pm$ 0.44 & 4.95 $\pm$ 0.07 & - \\
 & InstSpecHH + Closest & 1265.54 & 4.88 & - \\
 & InstSpecHH + LLM & 1264.78 $\pm$ 0.37 & 4.78 $\pm$ 0.01 & 1.02 \\
 & InstSpecHH + Classifier & \textbf{1262.81} & \textbf{4.62} & \textbf{-} \\
\midrule
\multirow{7}{*}{\centering CVRP} 
 & Closest Priority & 431.57 & 18.92 & - \\
 & EoH & 388.70 & 5.36 & - \\
 & ReEvo & 391.11 & 6.52 & - \\
\cmidrule(lr){2-5}
 & InstSpecHH w/o NS + LLM & 398.41 $\pm$ 0.56 & 8.89 $\pm$ 0.20 & 1.01 \\
 & InstSpecHH + Random & 385.70 $\pm$ 0.10 & 4.69 $\pm$ 0.02 & - \\
 & InstSpecHH + Closest & 384.46 & 4.40 & - \\
 & InstSpecHH + LLM & \textbf{384.34 $\pm$ 0.11} & \textbf{4.38} $\pm$ 0.02 & 1.01 \\
 & InstSpecHH + Classifier & 384.42 & 4.39 & - \\
\bottomrule
\end{tabular}
\vspace{4pt}
\end{table*}

\subsection{Performance Comparison}
\subsubsection{Intra-Subclass Generalization}
To answer \textit{RQ1} (whether instance-specific heuristic design can achieve better performance by taking instance heterogeneity into account), this section compares ISHH, which follows the instance-specific paradigm, with EoH and ReEvo, which follow the problem-specific paradigm. The comparison is conducted on 3,150 OBPP and 472 CVRP subclasses, where test instances share feature combinations with those in the corresponding training sets.
As presented in Table~\ref{tab:intra_main}, InstSpecHH significantly outperforms the problem-specific baselines (EoH and ReEvo), reducing the average optimality gap by 6.26\% for OBPP and 0.66\% for CVRP. This confirms the advantage of the instance-specific paradigm over problem-specific algorithms. Interestingly, while EoH and ReEvo evolve logic similar to the robust Best Fit algorithm for OBPP and outperform Closest Priority for CVRP, InstSpecHH achieves further improvements by addressing instance heterogeneity. Additionally, integrating the neighborhood search strategy yields a further optimality gap reduction of 2.34\% (OBPP) and 3.27\% (CVRP), emphasizing the necessity of local optima escape mechanisms during heuristic evolution. Therefore, taking instance heterogeneity into account leads to better heuristic performance (for \textit{RQ1}).

\subsubsection{Inter-Subclass Generalization}\label{sec:inter_gen}
To further answer \textit{RQ1}, this section considers the scenario of unseen feature combinations, and the experimental results are shown in Table~\ref{tab:inter_main}. Crucially, unlike intra-subclass scenarios, addressing these out-of-distribution instances necessitates an explicit \textit{algorithm selection} phase to route new instances to the most suitable pre-generated heuristics. Equipped with the LLM-based selection strategy (InstSpecHH+LLM), the proposed framework significantly surpasses the single-heuristic baselines (EoH and ReEvo), improving the optimality gap by 6.06\% for OBPP and 0.98\% for CVRP, while maintaining highly efficient LLM queries (averaging 1.01). Consequently, these results demonstrate that the framework retains its performance advantages in inter-subclass scenarios (for \textit{RQ1}).

To answer \textit{RQ2}, four selection strategies are compared: Random, Closest (distance-based matching via Eq.~\ref{eq:dist}), LLM, and Classifier. When comparing the selection strategies, the LLM-based approach leverages its reasoning capabilities to better capture complex feature-algorithm relationships, outperforming the Closest baseline by 0.1\% on OBPP. For CVRP, the gain over Closest is a marginal 0.02\%, due to the smaller candidate pool. These results indicate that LLMs can effectively enhance algorithm selection performance. Furthermore, the Classifier strategy achieves performance comparable to the LLM (optimality gaps of 4.62\% for OBPP and 4.39\% for CVRP). Consequently, the results demonstrate that the LLM and the Classifier serve as competitive adaptive selection mechanisms, effectively routing unseen instances to ensure robust generalization beyond the training distributions (for \textit{RQ2}). To further investigate the performance of LLM-based and Classifier-based strategies under different numbers of candidate algorithms, Section~\ref{sec:sen_candidate} presents a comparative experiment.

\begin{table*}[t]
\centering
\caption{\textbf{Intra-Subclass and Inter-Subclass Analysis for the Single Individual:} Performance comparison between InstSpecHH and heuristics individually designed for each problem instance. The \textbf{Time} column corresponds to the \textbf{online time cost}.}
\label{tab:inv_analysis}
\renewcommand{\arraystretch}{1.2}
\setlength{\tabcolsep}{8pt}
\begin{minipage}[t]{0.48\textwidth}
\centering
\textbf{(a) Intra-Subclass}\\[4pt]
\begin{tabular}{l l c c}
\toprule
\textbf{Problem} & \textbf{Method} & \textbf{Obj. ($\downarrow$)} & \textbf{Time / s} \\
\midrule
\multirow{3}{*}{OBPP}
 & EoH (Individual) & 1126.12 & 10120.75 \\
 & InstSpecHH w/o NS & 1124.04 & $6.65\times10^{-5}$ \\
 & InstSpecHH & 1101.93 & $6.37\times10^{-5}$ \\
\midrule
\multirow{3}{*}{CVRP}
 & EoH (Individual) & 351.32 & 7274.99 \\
 & InstSpecHH w/o NS & 361.38 & $1.05\times10^{-5}$ \\
 & InstSpecHH & 348.07 & $2.94\times10^{-5}$ \\
\bottomrule
\end{tabular}
\end{minipage}
\hfill
\begin{minipage}[t]{0.48\textwidth}
\centering
\textbf{(b) Inter-Subclass}\\[4pt]
\begin{tabular}{l l c c}
\toprule
\textbf{Problem} & \textbf{Method} & \textbf{Obj. ($\downarrow$)} & \textbf{Time / s} \\
\midrule
\multirow{3}{*}{OBPP}
 & EoH (Individual) & 1521.52 & 9231.41 \\
 & InstSpecHH w/o NS & 1537.20 & 6.57 \\
 & InstSpecHH & 1519.31 & 5.30 \\
\midrule
\multirow{3}{*}{CVRP}
 & EoH (Individual) & 417.52 & 8215.18 \\
 & InstSpecHH w/o NS & 419.80 & 2.95 \\
 & InstSpecHH & 409.74 & 4.75 \\
\bottomrule
\end{tabular}
\end{minipage}
\end{table*}

\subsection{Single Individual Analysis}
To answer \textit{RQ3} (how InstSpecHH compares with a strong tailored single instance baseline), we compare InstSpecHH with a highly specialized baseline, denoted as EoH (Individual), which constructs a customized heuristic algorithm for each individual test instance. Specifically, EoH (Individual) is trained using only the target instance itself, whereas these instances are not included in the training set of InstSpecHH. The evaluation is conducted on test instances sampled from 25 subclasses drawn from both the Intra- and Inter-Subclass distributions. The two methods are compared in terms of both solution quality and time cost. For a fair comparison, all methods use the DeepSeek-R1-Distill-Qwen-14B model.
\subsubsection{Performance on Individual Instances} 
As shown in Table~\ref{tab:inv_analysis}, InstSpecHH consistently outperforms the EoH (Individual) baseline. Under the Intra-Subclass setting, it reduces the objective value by 2.15\% (OBPP) and 0.93\% (CVRP). This advantage persists in the Inter-Subclass setting (LLM-based selector), with corresponding reductions of 0.15\% (OBPP) and 1.86\% (CVRP).
This superior performance is primarily driven by the Neighborhood Search (NS) strategy, which critically enables the generated heuristics to escape local optima. When the NS strategy is removed, EoH (Individual) generally outperforms InstSpecHH w/o NS. The heuristic construction process of InstSpecHH w/o NS is identical to that of EoH (Individual), except that it employs 30 training instances. Therefore, InstSpecHH achieves better performance than the single instance baseline (for \textit{RQ3}), due to the contribution of the NS strategy.

\begin{figure*}[htbp]
    \centering
    \subfloat[OBPP]{
        \includegraphics[width=0.42\textwidth]{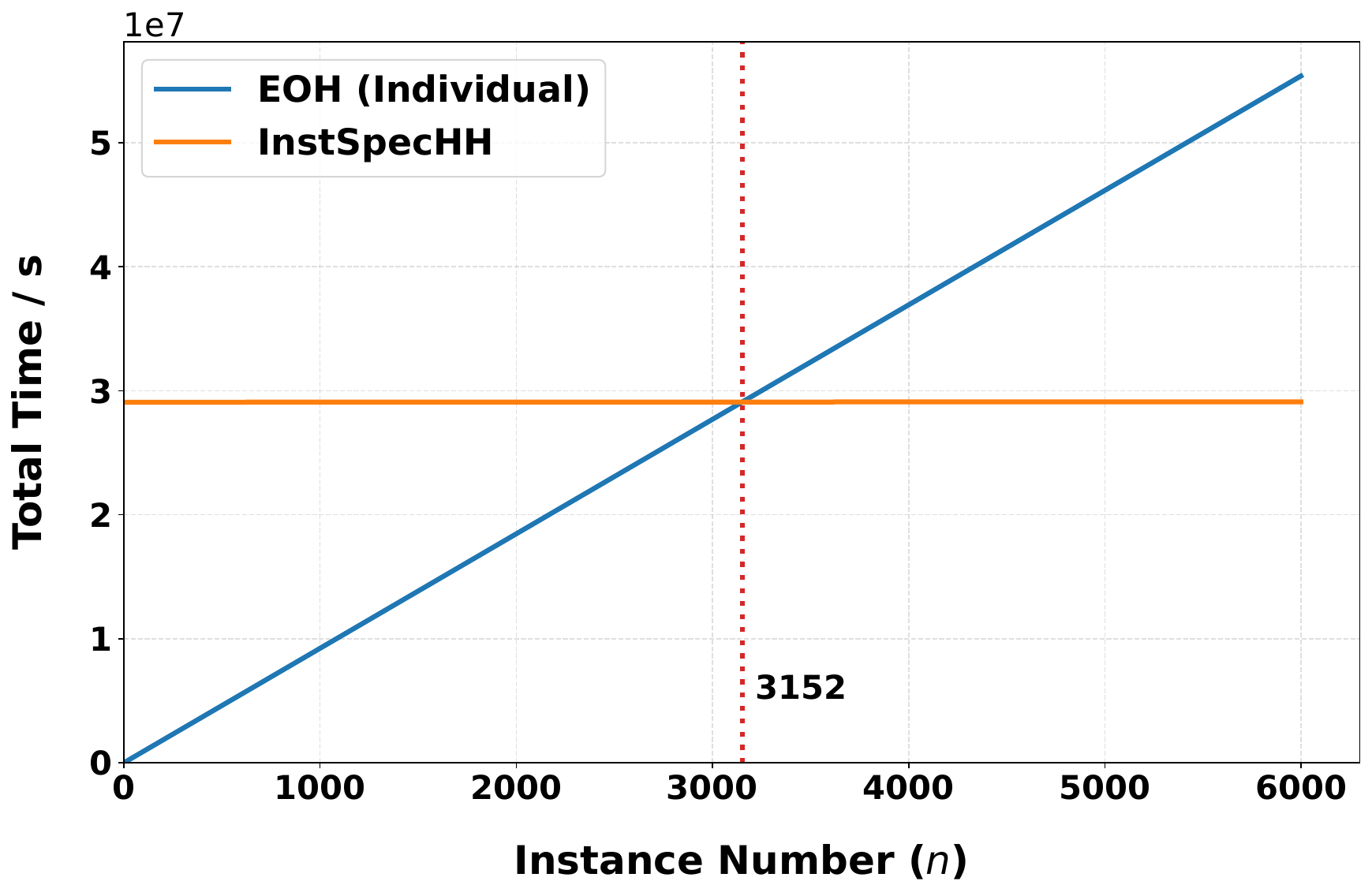}
        \label{fig:time_obpp}
    }
    \hspace{0.04\textwidth}
    \subfloat[CVRP]{
        \includegraphics[width=0.42\textwidth]{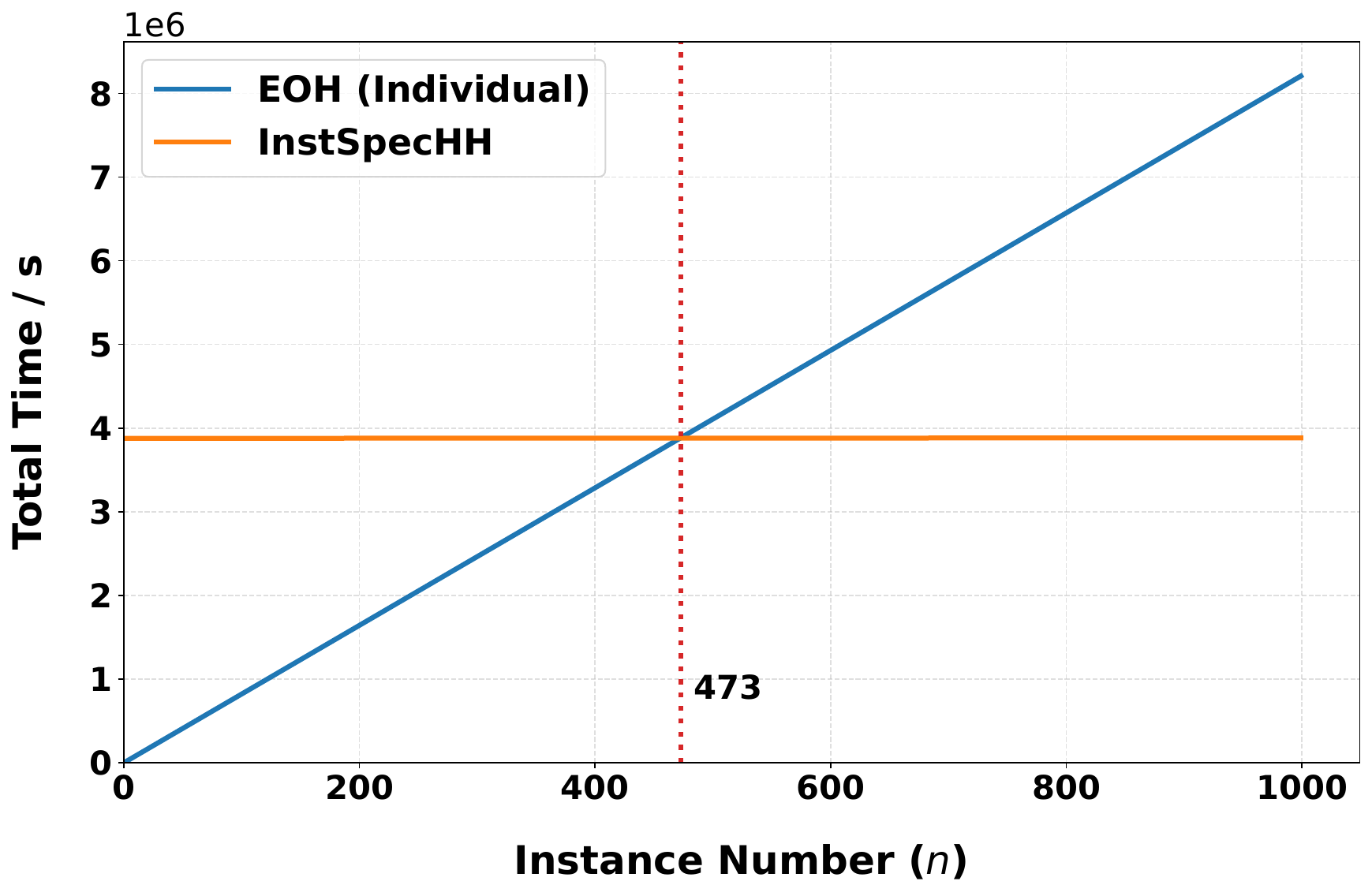}
        \label{fig:time_cvrp}
    }
    \caption{\textbf{Estimated Total Time Cost Analysis:} The total time cost is calculated as $T_{\text{total}} = T_{\text{offline}} + n \times T_{\text{online}}$. For EoH (Individual), $T_{\text{offline}} = 0$. For InstSpecHH, $T_{\text{offline}} = \#\text{intra-subclass} \times T_{\text{code design}}$, where $T_{\text{code design}}$ is estimated based on the online time cost of EoH (Individual). The online costs are estimated based on Table~\ref{tab:inv_analysis} (Inter-Subclass).}
    \label{fig:time_analysis}
    \vspace{4pt}
\end{figure*}

\subsubsection{Computational Efficiency Analysis}\label{sec:inv_efficiency}
The computational overhead is decomposed into an offline construction phase and an online inference phase. Although InstSpecHH incurs a relatively high offline cost to evolve subclass-specific algorithms, it drastically accelerates the online inference process. As demonstrated in Table~\ref{tab:inv_analysis}, the online inference time of InstSpecHH is approximately three orders of magnitude lower than the end-to-end evolutionary time required by EoH (Individual).

From the perspective of total time cost, which includes both offline (algorithm design) and online (algorithm selection) time costs, InstSpecHH demonstrates greater efficiency when solving a large number of problem instances, as shown in Figure~\ref{fig:time_analysis}. Due to the high offline design cost, the time costs of InstSpecHH and the online time cost of EoH (Individual) were estimated based on the data of Intra-Subclass in Table~\ref{tab:inv_analysis}. InstSpecHH front-loads its time cost during the design phase, resulting in a relatively high initial construction cost. However, as the number of problem instances increases, the amortized time cost per instance decreases substantially, leading to overall improved computational efficiency. For the OBPP problem, the break-even point occurs at around 3,152 instances, after which InstSpecHH becomes more time-efficient than EoH (Individual); for the CVRP problem, the crossover appears near 473 instances. These results demonstrate that InstSpecHH effectively reduces computational overhead and achieves better scalability when solving a large number of problem instances.

\begin{figure*}[htbp]
    \centering
    \subfloat[OBPP]{
        \includegraphics[width=0.42\textwidth]{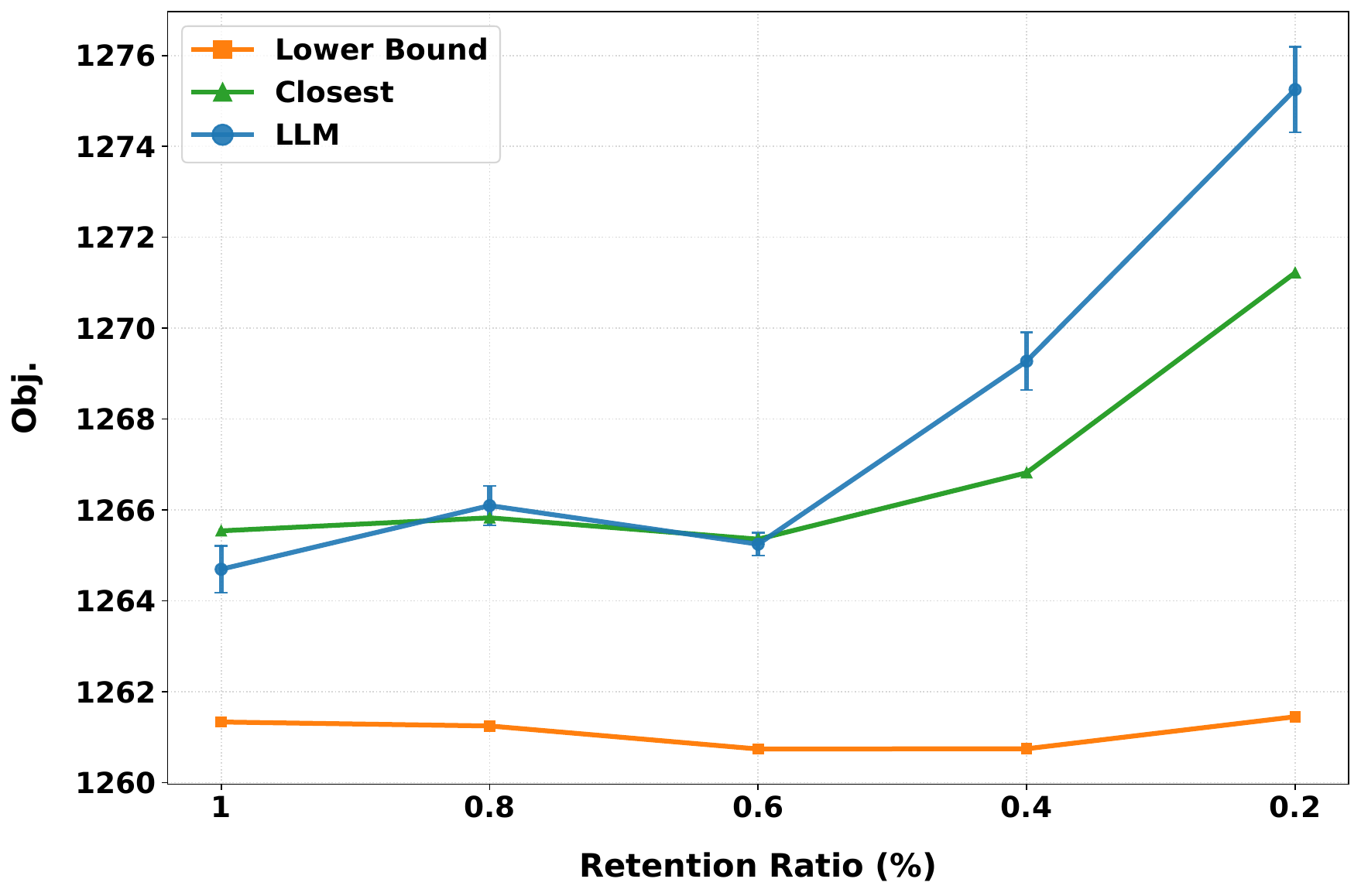}
        \label{fig:scal_obpp}
    }
    \hspace{0.04\textwidth}
    \subfloat[CVRP]{
        \includegraphics[width=0.42\textwidth]{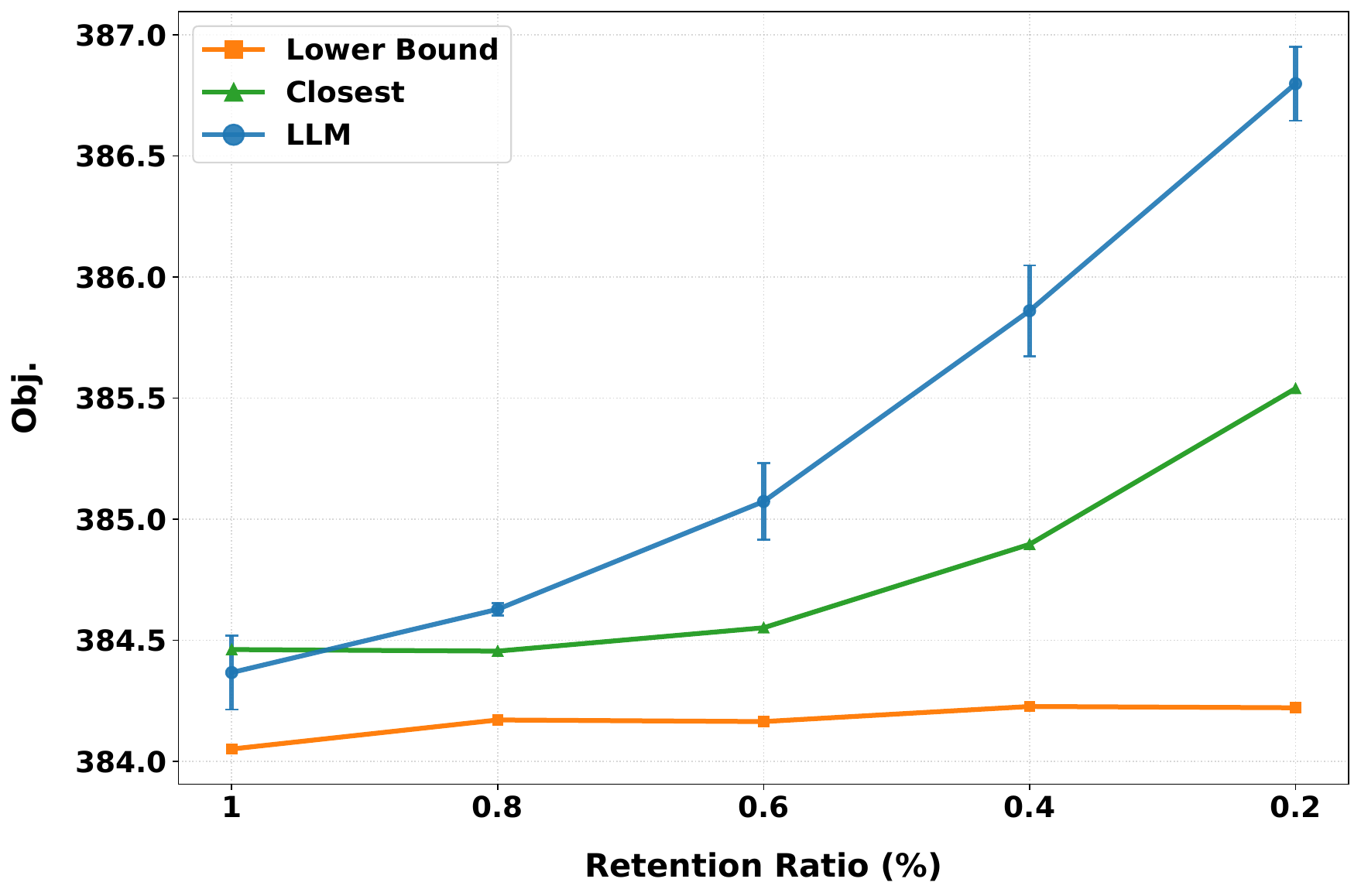}
        \label{fig:scal_cvrp}
    }
    \caption{\textbf{InstSpecHH Performance vs Number of Heuristic Algorithm}: Inter-subclass generalization performance under different heuristic retention ratios in InstSpecHH + LLM. Lower objective values indicate a higher quality of the selected heuristics.}
    \label{fig:scal}
    \vspace{4pt}
\end{figure*}

\subsection{Sensitivity Analysis}
\begin{figure*}[htbp]
    \centering
    \subfloat[OBPP]{
        \includegraphics[width=0.42\textwidth]{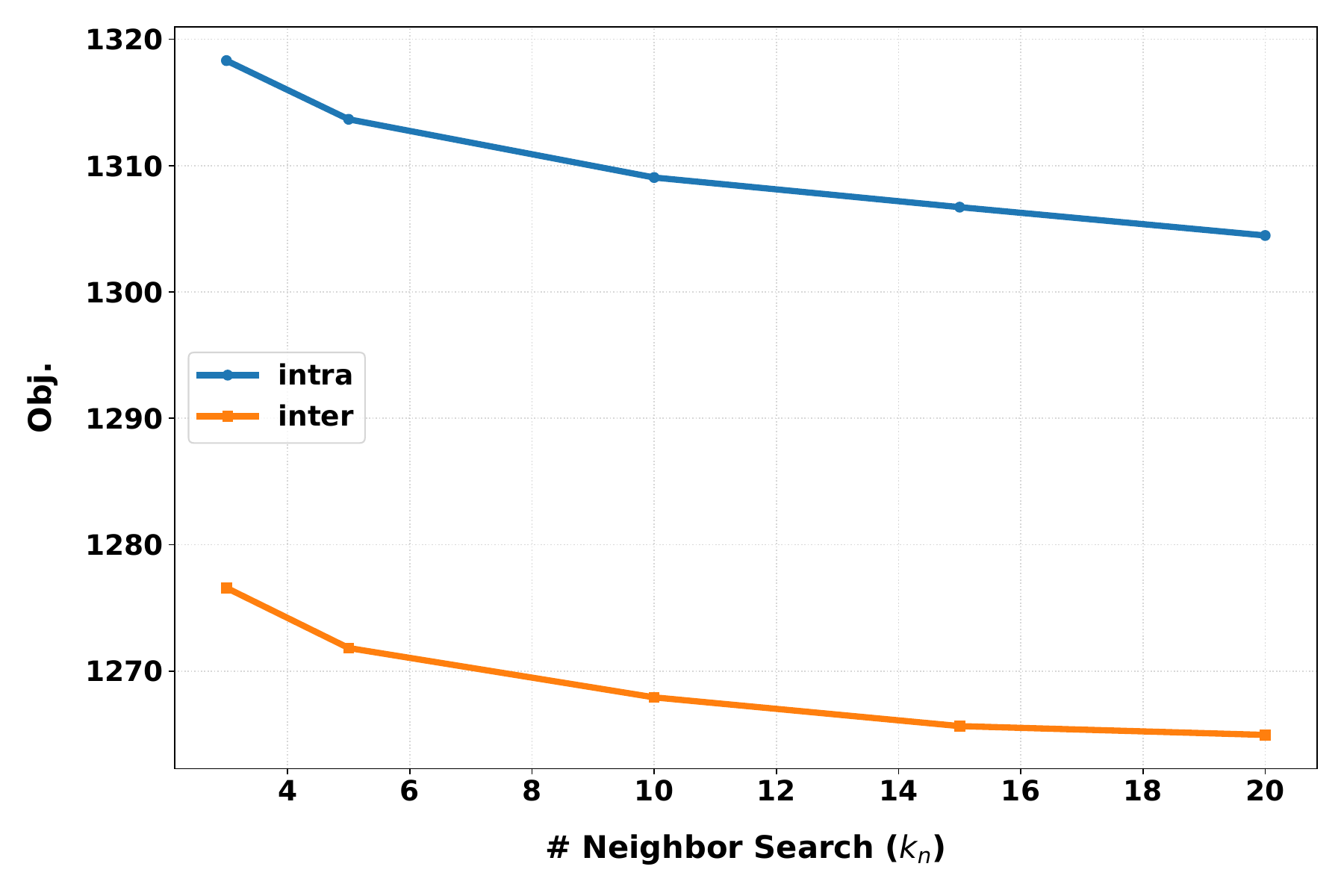}
        \label{fig:ns_obpp}
    }
    \hspace{0.04\textwidth}
    \subfloat[CVRP]{
        \includegraphics[width=0.42\textwidth]{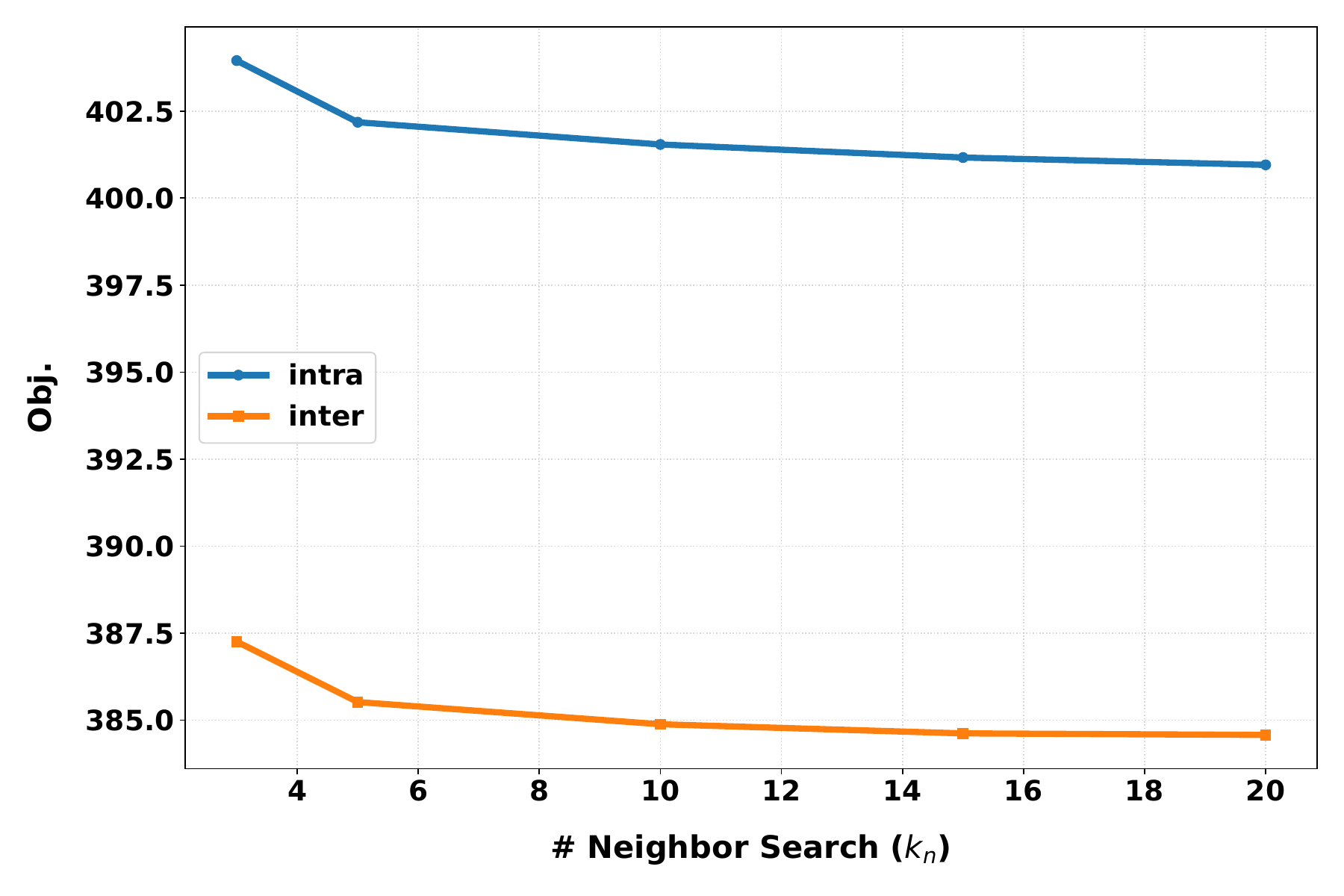}
        \label{fig:ns_cvrp}
    }
    \caption{\textbf{InstSpecHH Performance vs Number of Neighbor Search}: Comparison of the generalization performance of InstSpecHH + LLM under different neighborhood search sizes ($k_n$) in Intra- and Inter-Subclass.}
    \label{fig:ns_analysis}
    \vspace{4pt}
\end{figure*}

To answer \textit{RQ4} (i.e., how sensitive InstSpecHH is to its parameter settings), this section conducts a sensitivity analysis across three critical dimensions: the total number of heuristics algorithm pool, the NS size ($k_n$) during algorithm construction, and the candidate pool size ($k_c$) during algorithm selection.

\subsubsection{Sensitivity to the Algorithm Pool Size}
To assess the impact of the algorithm pool size on inter-subclass generalization, the framework is evaluated by randomly retaining 80\%, 60\%, 40\%, and 20\% of the generated heuristics. As illustrated in Figure~\ref{fig:scal}, performance degrades gradually with initial reductions, indicating a degree of beneficial redundancy within the algorithm pool. However, extreme reductions cause a sharp decline in the efficacy of both the LLM-based and Closest selection strategies. Furthermore, as the algorithm pool shrinks, the variance of the LLM-based strategy increases. This occurs because the widening gap between the target instance features and the available heuristic profiles impairs the LLM's reasoning accuracy. Conversely, the Lower Bound curve deteriorates relatively slowly, confirming the inherent robustness and broad adaptability of certain high-quality heuristics even within a severely constrained algorithm pool.

\subsubsection{Sensitivity to Neighbor Number in NS}
The influence of the neighborhood search size is evaluated by varying $k_n$ from 2 to 20 during algorithm construction, keeping all other parameters constant. Figure~\ref{fig:ns_analysis}  demonstrates that increasing $k_n$ consistently reduces the objective value across both OBPP and CVRP, as the NS strategy effectively enables the generated heuristics to escape local optima. Nevertheless, the marginal performance gains diminish as the neighborhood expands, nearly saturating at $k_n = 20$. To strike an optimal balance between evolutionary performance and computational efficiency, a neighborhood size of 20 is adopted for the InstSpecHH framework.
\subsubsection{Sensitivity to the Candidate Number}\label{sec:sen_candidate}
\begin{figure*}[htbp]
    \centering
    \subfloat[OBPP]{
        \includegraphics[width=0.42\textwidth]{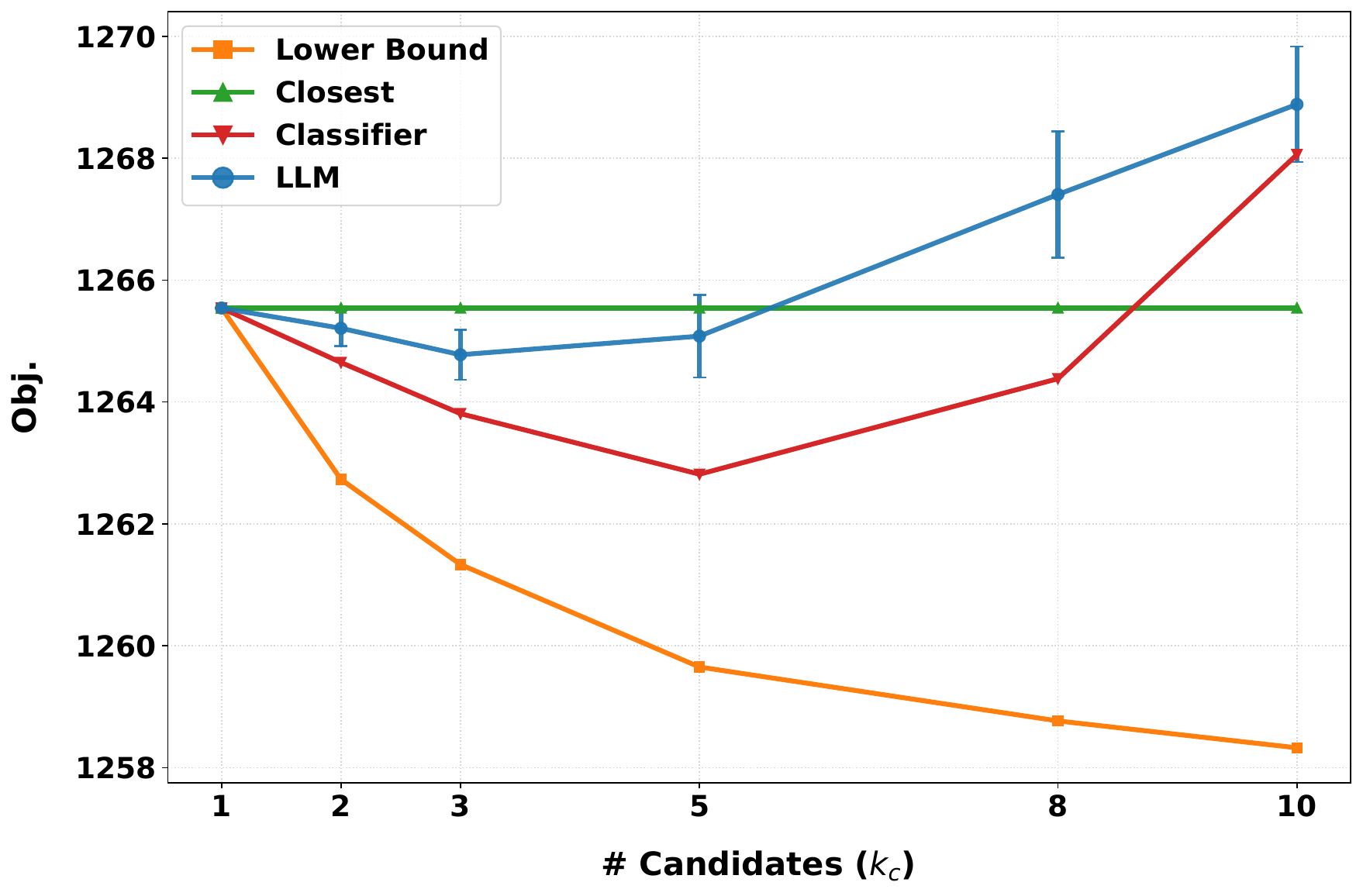}
        \label{fig:topk_obpp}
    }
    \hspace{0.04\textwidth}
    \subfloat[CVRP]{
        \includegraphics[width=0.42\textwidth]{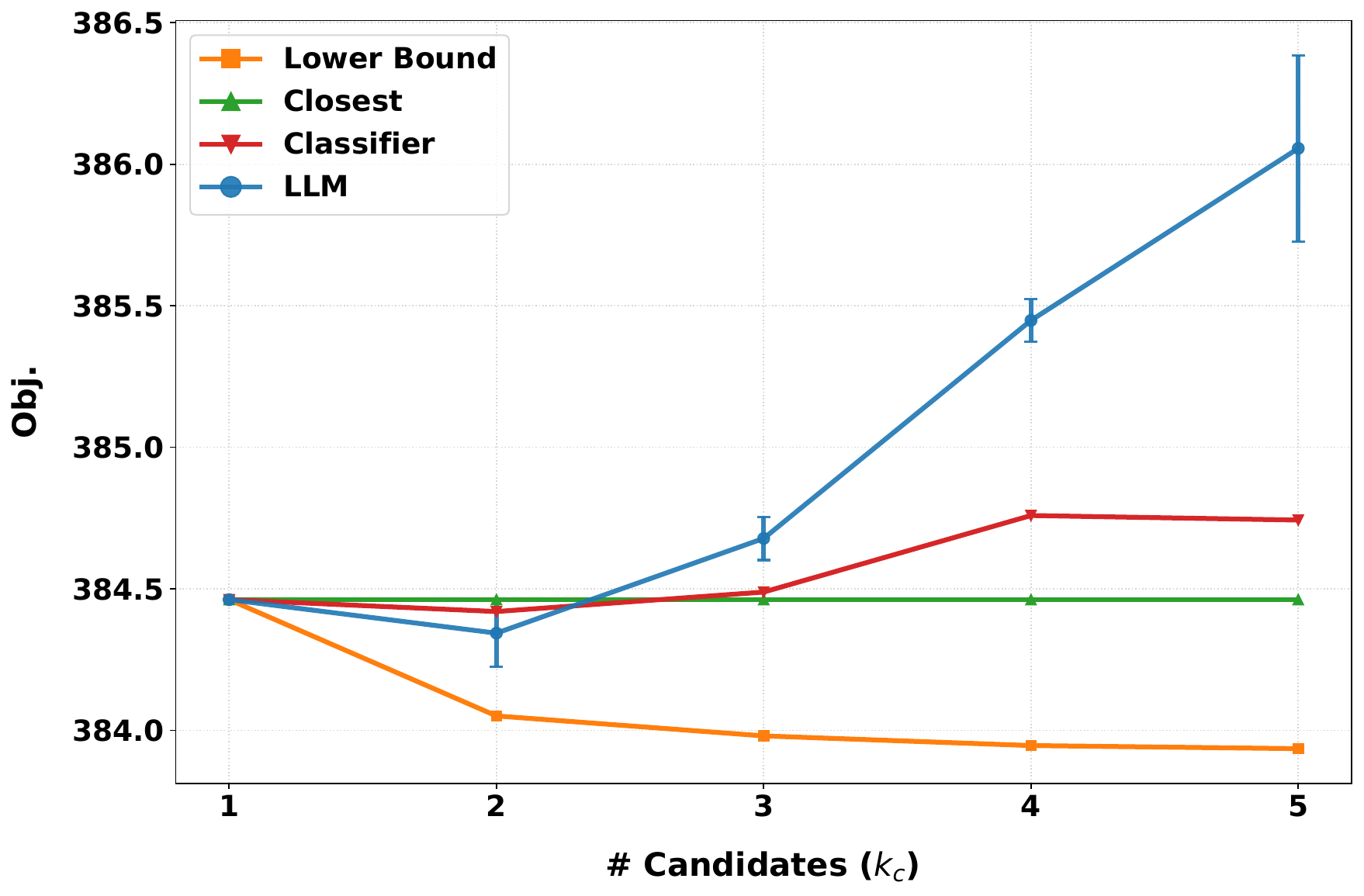}
        \label{fig:topk_cvrp}
    }
    \caption{\textbf{InstSpecHH Performance vs Number of Candidates}: Inter-subclass generalization performance comparison of different algorithm selection strategies used by InstSpecHH under varying numbers of candidate heuristics ($k_c$). Lower objective value (Obj.) indicates a higher quality of the selected heuristic.}
    \label{fig:topk}
    \vspace{4pt}
\end{figure*}

Figure~\ref{fig:topk} illustrates the impact of the candidate algorithm number ($k_c$) on the algorithm selection phase. Naturally, the Lower Bound continuously improves as a larger $k_c$ incorporates more potentially optimal heuristics. The Closest strategy remains unaffected, as the nearest neighbor does not change with a larger retrieval pool. For the LLM-based strategy, performance initially improves but subsequently declines as $k_c$ increases further. An overly large candidate set introduces irrelevant heuristics, effectively acting as noise that impairs the LLM's ability to pinpoint the optimal choice, thereby increasing solution variance. In contrast, the Classifier-based selection demonstrates highly stable performance across varying $k_c$ values. Because the neural network directly maps instance features to heuristic identifiers, it remains robust against candidate pool expansion, maintaining a consistent level of solution quality. Therefore, with a large number of candidate algorithms, the classifier provides a more robust selection mechanism than the LLM.

\subsection{Feature Analysis}\label{sec:feature_analysis}
\begin{figure*}[htbp]
    \centering
    \includegraphics[width=0.95\textwidth]{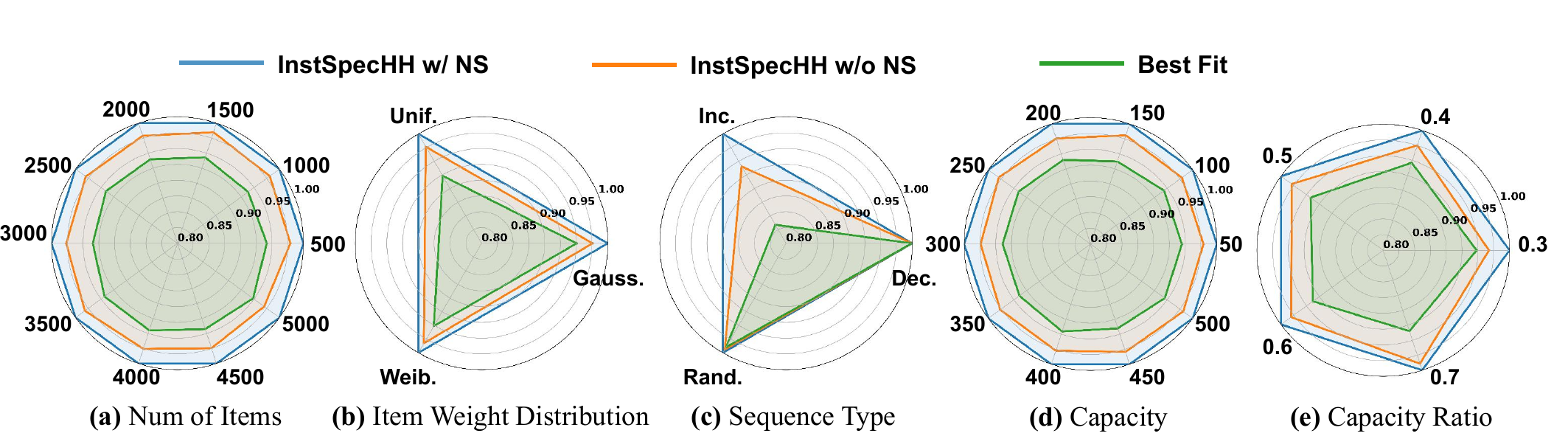}
    \caption{\textbf{OBPP Feature Analysis}: Performance comparison of InstSpecHH and individual heuristics on OBPP across varying problem features. The evaluation metric is the 1-Opt. gap, where values closer to 1 indicate higher algorithm quality.}
    \label{fig:radar_obpp}
    \vspace{6pt}
\end{figure*}

\begin{figure*}[htbp]
    \centering
    \includegraphics[width=0.95\textwidth]{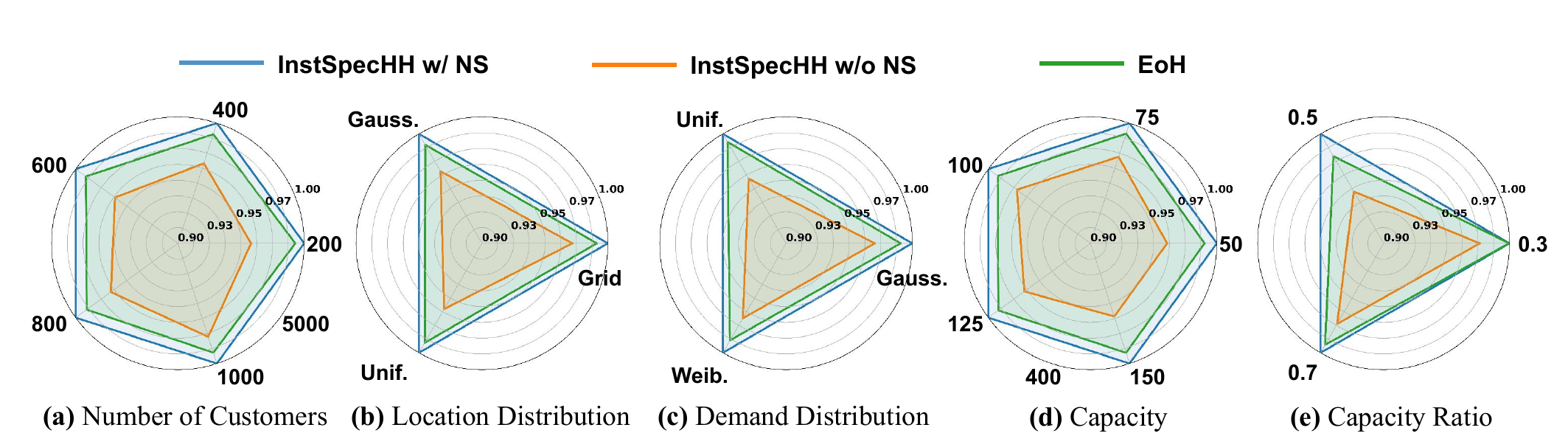}
    \caption{\textbf{CVRP Feature Analysis}: Performance comparison of InstSpecHH and individual heuristics on CVRP across varying problem features. The evaluation metric is the 1-Opt. gap, where values closer to 1 indicate higher algorithm quality.}
    \label{fig:radar_cvrp}
    \vspace{6pt}
\end{figure*}

To answer \textit{RQ5} (under which instance feature characteristics the instance-specific paradigm outperforms the problem-specific paradigm), we analyze the performance variations of InstSpecHH across different problem feature dimensions. Both InstSpecHH variants, with and without the Neighborhood Search component, are compared against baseline algorithms using $1 - \text{Opt. gap}$ as the evaluation metric.
The experimental results are shown in Figure~\ref{fig:radar_obpp} and~\ref{fig:radar_cvrp}.

For the OBPP, InstSpecHH demonstrates substantial gains under specific configurations: a notable 16.61\% improvement over the Best Fit algorithm for increasing sequence types, and a 7.66\% improvement under Uniform item weight distributions (For \textit{RQ5}). The analysis further isolates the critical role of NS across varying difficulty levels. Regarding capacity ratios, InstSpecHH achieves optimal gains at a ratio of 0.7 (6.53\%) and minimal gains at 0.3 (5.20\%). In contrast, without NS, these improvements drop to 1.10\% and 3.24\%, respectively, suggesting that lower capacity ratios increase the risk of local optima. Furthermore, while InstSpecHH maintains stable superiority across different scales, without NS, its advantage over Best Fit shrinks from 4.26\% (at 1000 items) to a mere 2.19\% (at 5000 items). This degradation clearly indicates a higher likelihood of suboptimal behavior as the instance scale grows. Finally, with respect to the bin capacity, performance gains remain relatively uniform, fluctuating narrowly between 5.46\% and 6.24\%.

For the CVRP, InstSpecHH consistently outperforms the EoH baseline, yielding an average improvement of 0.66\%. This advantage becomes particularly pronounced under specific configurations, peaking at a 2.1\% improvement for a capacity ratio of 0.5 and a 1.14\% gain for Weibull-distributed demands (for \textit{RQ5}). Notably, ablating the NS strategy degrades performance to approximately 2.5\% below the EoH baseline. This stark contrast demonstrates that the instance-specific evolutionary process in CVRP scenarios is highly susceptible to local optima, which could be effectively mitigated by the NS component.

\subsection{Model Analysis}
\begin{figure*}[htbp]
    \centering
    \subfloat[Algorithm Design]{
        \includegraphics[width=0.42\textwidth]{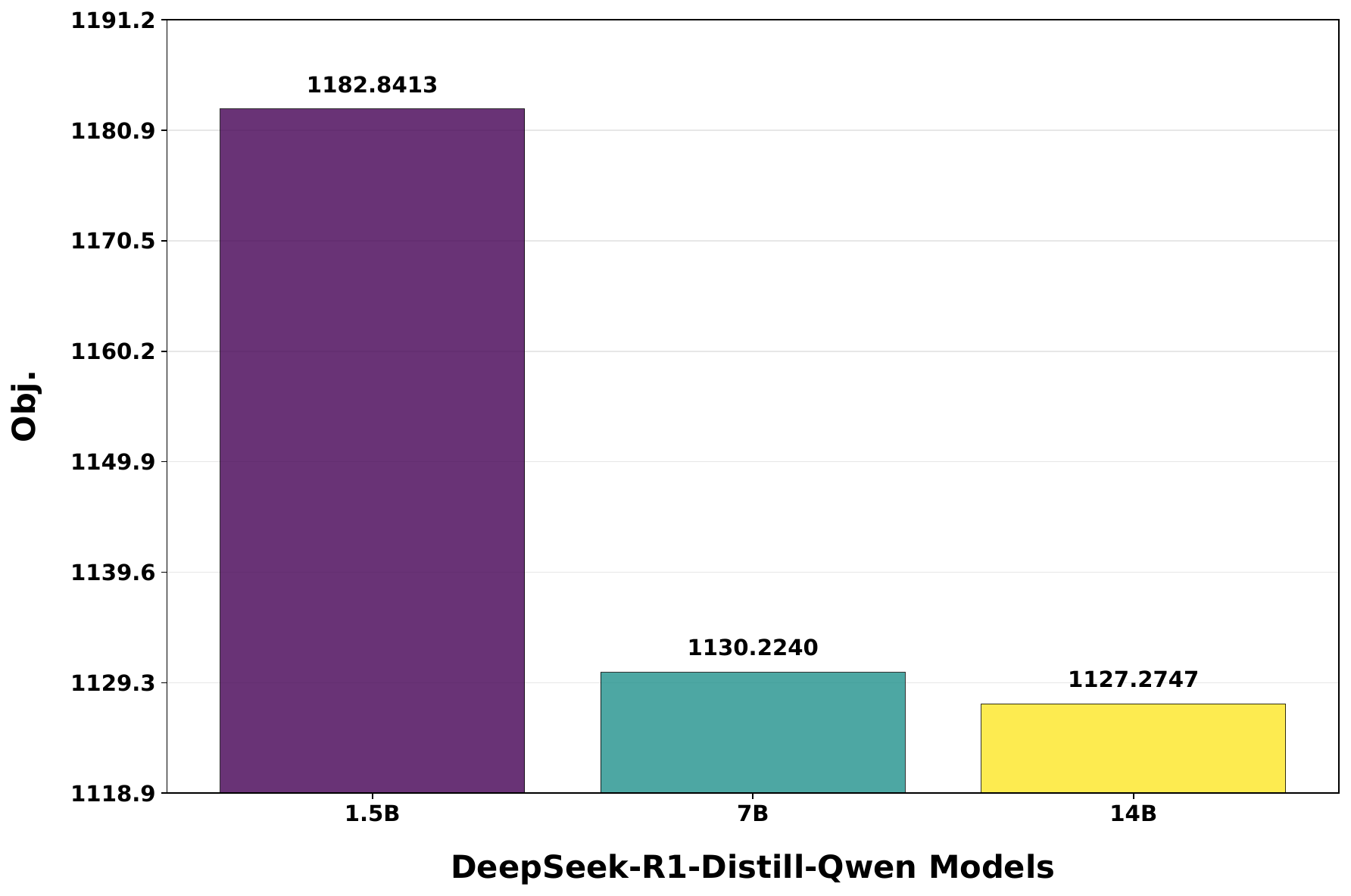}
        \label{fig:model_generation}
    }
    \hspace{0.04\textwidth}
    \subfloat[Algorithm Selection]{
        \includegraphics[width=0.42\textwidth]{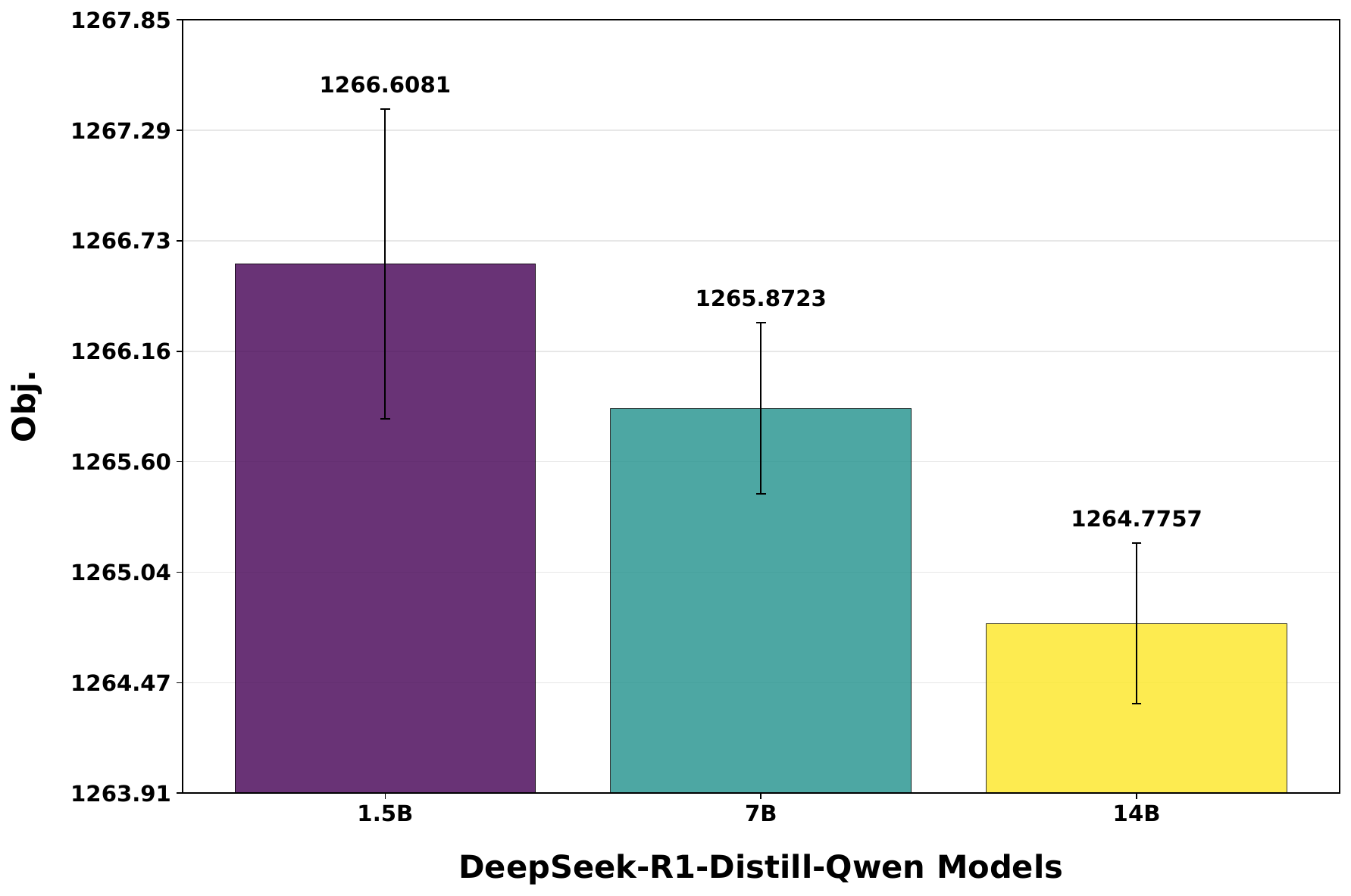}
        \label{fig:model_selection}
    }
    \caption{\textbf{Performance of Different LLM Models on OBPP for Algorithm Design and Selection.}}
    \label{fig:model_analysis_obpp}
    \vspace{4pt}
\end{figure*}

To examine the impact of different LLMs on both heuristic generation and selection, we compare the DeepSeek-R1-Distill-Qwen 1.5B, 7B, and 14B variants on the OBPP. In the heuristic generation phase, 25 problem subclasses are randomly selected for evolutionary construction due to the high computational cost. In the heuristic selection phase, the number of candidate heuristics is fixed at $k_c = 3$ for consistency.

As shown in Figure~\ref{fig:model_analysis_obpp}, model size substantially affects both generation and selection performance. Larger models generally exhibit stronger reasoning and generalization abilities, leading to better overall results. In heuristic generation (Figure~\ref{fig:model_generation}), the 7B and 14B models outperform the 1.5B model, achieving objective values of 1130.22 and 1127.27, respectively, compared with 1182.84. In heuristic selection (Figure~\ref{fig:model_selection}), performance also improves slightly with model size, with the objective value decreasing from 1266.61 (1.5B) to 1264.78 (14B). In addition, the 7B and 14B models show lower variance, indicating better stability.

Overall, model capacity has a clear impact on both heuristic generation and selection. Considering the trade-off between performance and inference cost, we adopt DeepSeek-R1-Distill-Qwen 14B as the default LLM.

\section{Conclusion and Dissusion}\label{sec:conclusion}
This paper introduces InstSpecHH, a framework leveraging LLMs to automatically generate and select instance-specific heuristics tailored to distinct problem subclasses. Extensive evaluations on the OBPP and CVRP demonstrate that InstSpecHH significantly outperforms state-of-the-art baselines (EoH and ReEvo) across both known distributions and previously unseen instances. Crucially, to handle out-of-distribution scenarios, LLM-based algorithm selection is compared against the traditional neural network classifier. Both paradigms achieve highly competitive selection quality, showing the viability of LLMs for automated heuristic selection. Furthermore, a persistent gap between current selection performance and the theoretical lower bound underscores significant untapped potential for future advancements in optimal algorithm selection. Ultimately, InstSpecHH establishes a robust, highly adaptable paradigm for automated, instance-aware heuristic design in combinatorial optimization.

The potential challenges in applying InstSpecHH to other optimization domains largely depend on the design of problem features. Fortunately, many well-studied optimization domains already provide a rich collection of representative features~\cite{Hutter_2014} that can be directly incorporated into InstSpecHH. For less-explored domains, more generic methodologies for feature extraction and relevance analysis, such as landscape analysis~\cite{Pitzer2012} or LLMs, can be employed to identify key instance features that influence heuristic performance. Once the problem features and the objective function are well established, InstSpecHH can be naturally extended beyond traditional combinatorial problems.

Another limitation of InstSpecHH lies in its relatively high offline computational cost. The construction of the heuristic library requires multiple rounds of heuristic generation and evaluation, leading to substantial LLM query overhead. A promising avenue for future work is to reuse or adapt heuristics generated for existing subclasses to accelerate heuristic construction for new subclasses, thereby reducing overall computation while preserving performance. Such transfer-based mechanisms would further enhance the practicality and scalability of the InstSpecHH framework across diverse optimization domains.




\bibliographystyle{IEEEtran}
\bibliography{main}

\newpage
\markboth{Supplementary Material}{Supplementary Material}
\begin{figure*}[!t]
\centering
{\Large\bfseries Supplementary Material For \\[0.8em]
``LLM-Driven Instance-Specific Heuristic Generation and Selection''}
\vspace{0.5em}
\end{figure*}

\section{Candidate Algorithm Description Template}
\begin{figure}[t]
    \centering
    \includegraphics[width=0.4\textwidth]{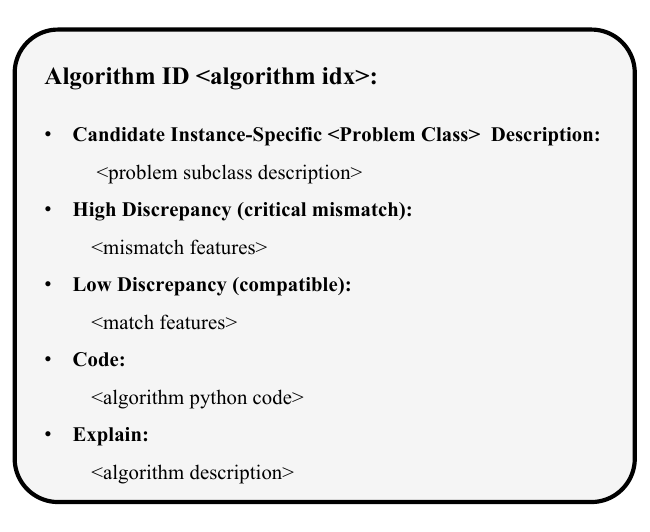}
    \vskip -0.1in
    \caption{Candidate Algorithm Description Template.}
    \label{fig:algo_temp_prompt}
\end{figure}

To enable reliable reasoning and selection, each candidate heuristic is represented using a structured template that describes the corresponding algorithm, as shown in Fig.~\ref{fig:algo_temp_prompt}. This template is designed to help the LLM evaluate the relationship between each candidate algorithm and the target instance, thereby facilitating effective heuristic selection.

\section{Heuristic Algorithm Generation Prompt}~\label{sec:gene_prompt}
\begin{figure}[t]
    \centering
    \includegraphics[width=0.4\textwidth]{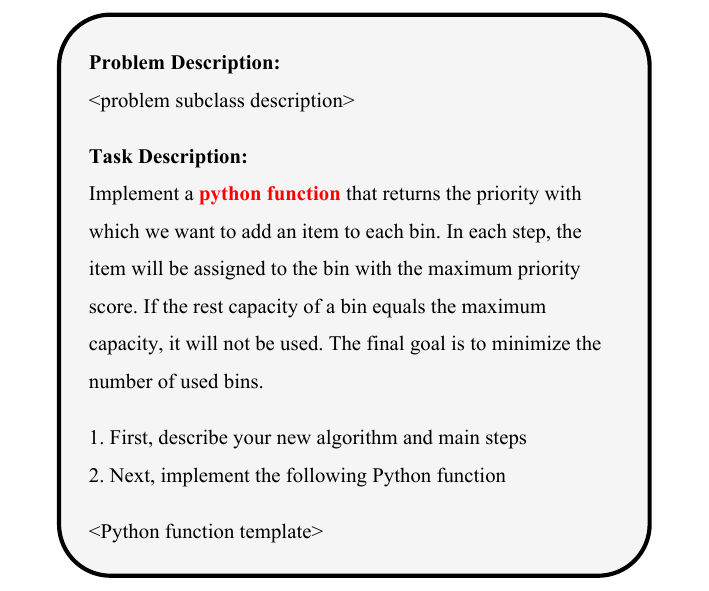}
    \vskip -0.1in
    \caption{Heuristic Algorithm Design Prompt Template for OBPP.}
    \label{fig:algo_generation_prompt}
\end{figure}

\begin{figure*}[t]
    \centering
    \includegraphics[width=\textwidth]{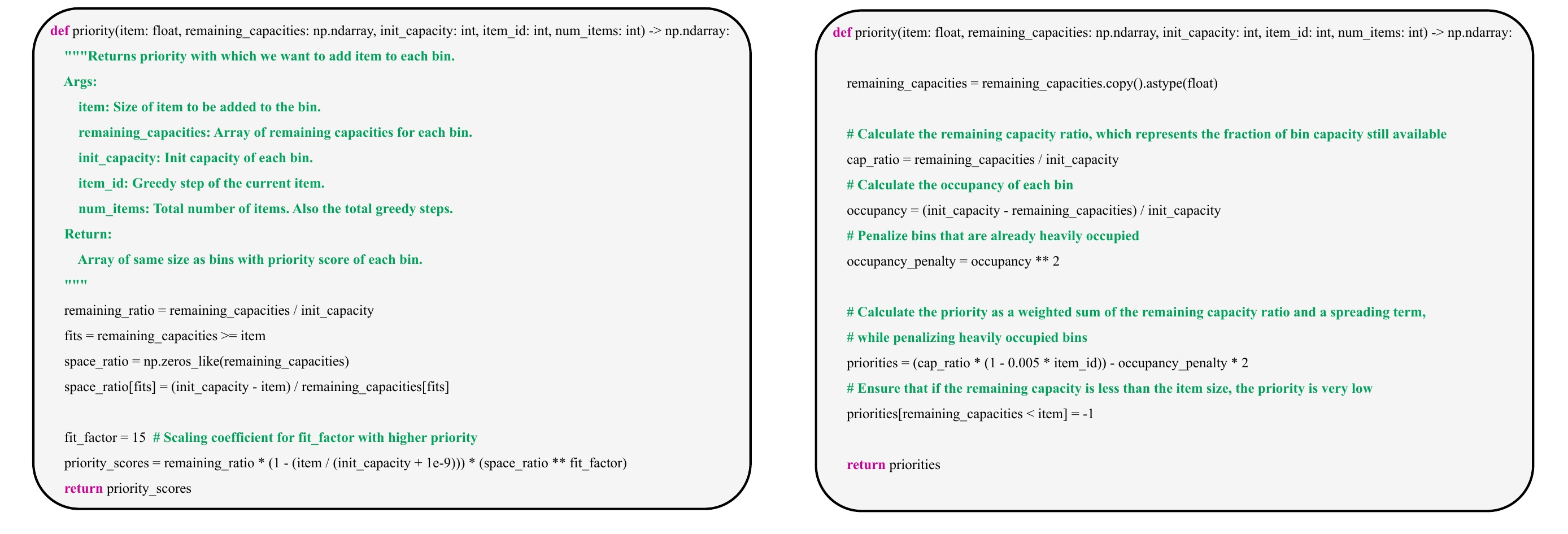}
    \vskip -0.1in
    \caption{Python Function Example for OBPP Heuristic Algorithm.}
    \label{fig:obpp_code}
\end{figure*}

\begin{figure*}[t]
    \centering
    \includegraphics[width=\textwidth]{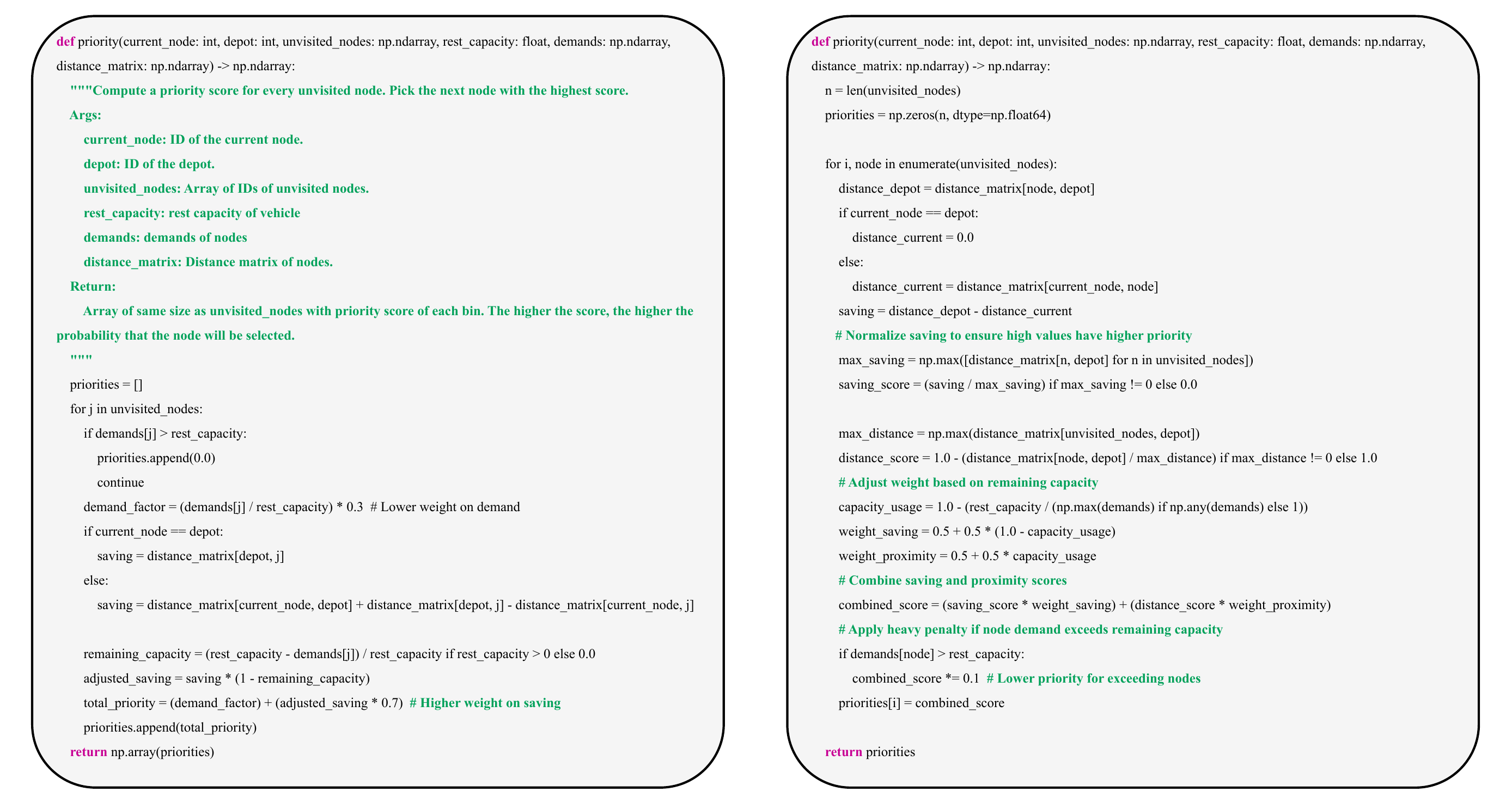}
    \vskip -0.1in
    \caption{Python Function Example for CVRP Heuristic Algorithm.}
    \label{fig:cvrp_code}
\end{figure*}

Figure~\ref{fig:algo_generation_prompt} presents the prompt used for heuristic function generation, while Figures~\ref{fig:obpp_code} and~\ref{fig:cvrp_code} show example Python implementations of heuristic functions for OBPP and CVRP, respectively.

\section{Problem Feature Description}
\begin{figure*}[htbp]
    \centering
    \includegraphics[width=0.75\textwidth]{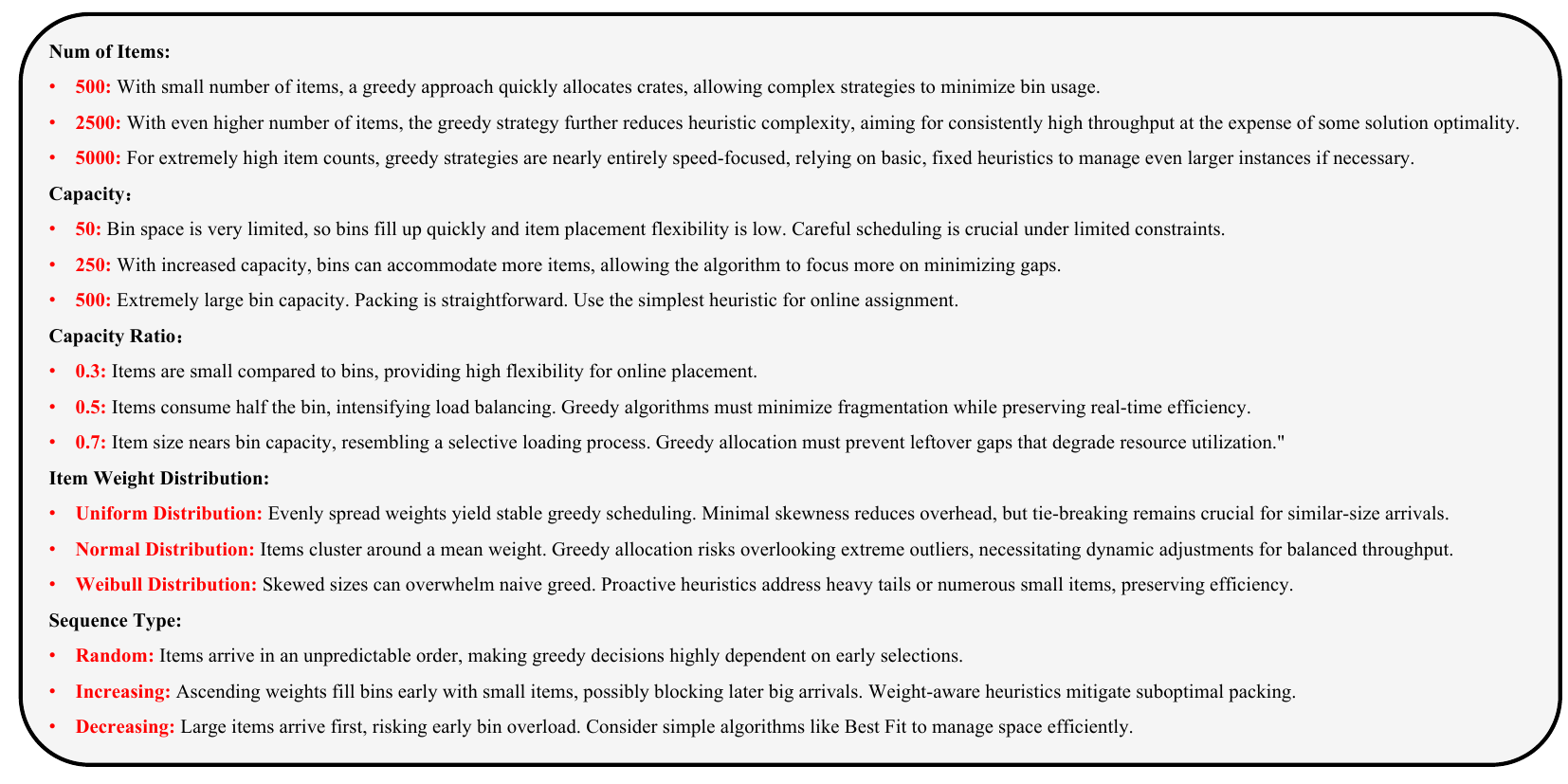}
    \vskip -0.1in
    \caption{OBPP Feature Description.}
    \label{fig:algo_desc_prompt_obpp}
\end{figure*}

\begin{figure*}[htbp]
    \centering
    \includegraphics[width=0.75\textwidth]{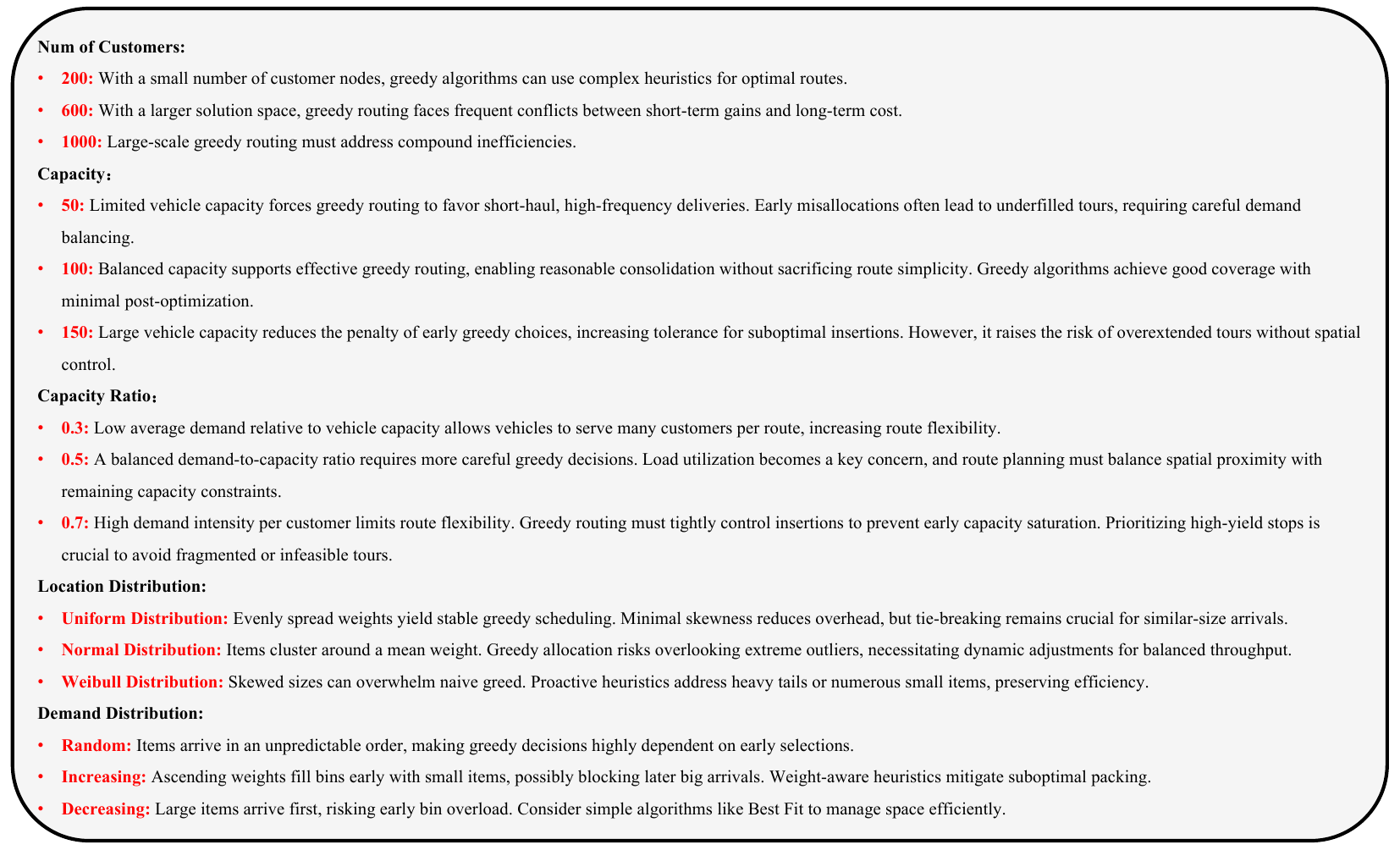}
    \vskip -0.1in
    \caption{CVRP Feature Description.}
    \label{fig:algo_desc_prompt_cvrp}
\end{figure*}

In this section, detailed descriptions of the constructed features for both the Online Bin Packing Problem (OBPP) and the Capacitated Vehicle Routing Problem (CVRP) are provided in Figures~\ref{fig:algo_desc_prompt_obpp} and~\ref{fig:algo_desc_prompt_cvrp}, respectively. These features are designed to capture the key factors that influence heuristic performance from multiple perspectives, including problem scale, capacity constraints, and distributional features.

\section{Evolutionary Operator Pool}\label{sec:evo_ope_pool}
This work utilizes the evolutionary operations provided by the EoH~\cite{fei2024eoh} to generate heuristic algorithms. Our framework, InstSpecHH, can be seamlessly integrated with other algorithm design approaches such as FunSearch~\cite{romera2024funsearch} and ReEvo~\cite{haoran2024reevo}. This section primarily introduces the EoH Evolutionary Operator Pool, which consists of two types of operators: crossover and mutation

The crossover operators combine two different parent heuristics to generate a new child algorithm. Two types of crossover operators are employed:
\begin{itemize}
    \item \textbf{E1:} Produce child heuristics that differ substantially from their parent heuristics.
    \item \textbf{E2:} Produce child heuristics that retain the essential ideas of their parents.
\end{itemize}

The mutation operators modify individual heuristics to explore unexplored regions of the search space. Three types of mutation operations are included:
\begin{itemize}
    \item \textbf{M1:} Generate an improved child heuristic by refining and adjusting the structural or functional design of the parent heuristic.
    \item \textbf{M2:} Produce an enhanced child heuristic through parameter tuning of the parent heuristic to achieve better local optimization.
    \item \textbf{M3:} Create a simplified child heuristic by removing redundant components from the parent heuristic.
\end{itemize}

Together, these evolutionary operations enable a balanced \emph{exploration-exploitation} process that continuously improves the quality and diversity of generated heuristic algorithms.


 




\vfill

\end{document}